%% file: main.tex
\documentclass[10pt,twocolumn,letterpaper]{article}

\usepackage[pagenumbers]{cvpr}

\input{preamble}
\usepackage{float}

\definecolor{cvprblue}{rgb}{0.21,0.49,0.74}
\usepackage[pagebackref,breaklinks,colorlinks,allcolors=cvprblue]{hyperref}

\def\confName{CVPR}
\def\confYear{2027}

\title{BayesPath: Reliability-Aware Bayesian Reasoning\\for Whole-Slide Pathology Agents}

\author{
Bryan Wong$^{1,2,\dagger}$\quad
Xun Xu$^{2,3,*}$\quad
Jongwoo Kim$^{1}$\quad
Huazhu Fu$^{2}$\quad
Nancy F. Chen$^{2,3}$\quad
Mun Yong Yi$^{1,*}$\\[0.4em]
$^{1}$Korea Advanced Institute of Science and Technology (KAIST)\\
$^{2}$Institute of Advanced Intelligence and Computing (IAIC), A*STAR\\
$^{3}$Centre for Frontier AI Research (CFAR), A*STAR\\[0.3em]
{\tt\small bryan.wong@kaist.ac.kr, xu\_xun@a-star.edu.sg, munyi@kaist.ac.kr}
}

\newcommand{\tabno}[1]{\setcounter{table}{\numexpr#1-1\relax}}

\begin{document}
\raggedbottom
\twocolumn[{%
\renewcommand\twocolumn[1][]{#1}%
\maketitle
\vspace{-2.2em}
\begin{center}
  {\setlength{\unitlength}{\textwidth}%
  \begin{picture}(1,0.33749)
    \put(0,0){\includegraphics[width=\textwidth]{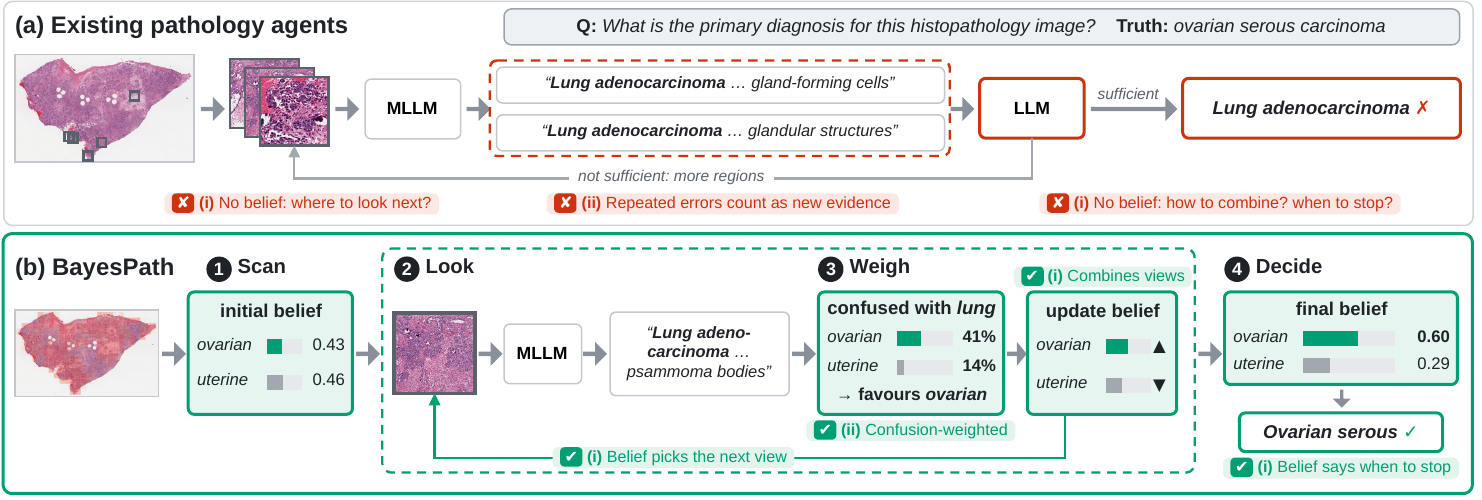}}%
    \put(0.13725,0.14356){\textcolor{white}{\rule{0.05943\textwidth}{0.02219\textwidth}}}%
    \put(0.17759,0.14356){\includegraphics[trim=97.60 102.09 571.26 122.13,clip,width=0.05943\textwidth]{figs/bayespath_fig1.pdf}}%
    \put(0.12216,0.15159){\textcolor[HTML]{1F2328}{\usefont{OT1}{phv}{b}{n}\fontsize{8}{8}\selectfont(Ours)}}%
  \end{picture}}%
  \vspace{-0.6em}
  \captionsetup{hypcap=false}
  \captionof{figure}{\textbf{Motivation and overview of \method.} The MLLM repeatedly misreads ovarian carcinoma as lung adenocarcinoma. (a)~Existing agents lack a belief to pick views, combine them or stop~(i) and count repeated errors as new evidence~(ii), so they answer lung. (b)~\method's belief picks views, combines them by Bayes' rule and says when to stop~(i); weighted by confusions, ``lung'' readings support ovarian~(ii).}
  \label{fig:teaser}
\end{center}
\vspace{0.6em}
}]
\begin{NoHyper}\let\thefootnote\relax\footnotetext{$^{\dagger}$Work done during a research attachment at A*STAR.}\footnotetext{$^{*}$Corresponding authors.}\end{NoHyper}

\input{sec/0_abstract}
\input{sec/1_intro}
\input{sec/2_related}
\input{sec/3_method}
\input{sec/4_experiments}

\input{sec/5_conclusion}

\paragraph{Acknowledgements.}
This research/project is supported by the National Research Foundation, Singapore under its National Large Language Models Funding Initiative (AISG Award No: AISG-NMLP-2024-004). Any opinions, findings and conclusions or recommendations expressed in this material are those of the author(s) and do not reflect the views of National Research Foundation, Singapore.

\input{sec/6_llm_statement}
{
    \small
    \bibliographystyle{ieeenat_fullname}
    \bibliography{main}
}

\clearpage
\setcounter{figure}{0}\setcounter{table}{0}\setcounter{equation}{0}\setcounter{algorithm}{0}
\renewcommand{\thefigure}{S\arabic{figure}}
\renewcommand{\thetable}{S\arabic{table}}
\renewcommand{\theequation}{S\arabic{equation}}
\renewcommand{\thealgorithm}{S\arabic{algorithm}}
\setcounter{topnumber}{4}\setcounter{dbltopnumber}{3}\setcounter{totalnumber}{6}
\renewcommand{\topfraction}{0.9}\renewcommand{\dbltopfraction}{0.9}\renewcommand{\textfraction}{0.07}
\renewcommand{\floatpagefraction}{0.85}\renewcommand{\dblfloatpagefraction}{0.85}
\makeatletter\setlength{\@fptop}{0pt}\setlength{\@fpsep}{\floatsep}\setlength{\@fpbot}{0pt plus 1fil}\makeatother
\input{sec/X_suppl}

\end{document}

%% file: preamble.tex
\usepackage{graphicx}
\usepackage{booktabs}
\usepackage{multirow}
\usepackage{pifont}
\usepackage{amsmath,amssymb}
\usepackage{xcolor}
\usepackage{colortbl}
\usepackage{tikz}
\usetikzlibrary{arrows.meta,positioning,fit,backgrounds,calc,shapes.misc,shadows,decorations.pathreplacing,shapes.geometric}
\usepackage{caption}
\usepackage{xspace}
\usepackage{enumitem}
\usepackage{algorithm}
\usepackage{algpseudocode}
\usepackage{microtype}
\newcommand{\method}{BayesPath\xspace}

\newcommand{\pms}[2]{#1{\scriptsize$\pm$#2}}     %
\DeclareMathOperator*{\argmax}{arg\,max}
\newcommand{\qbar}{\bar q}
\definecolor{cbblue}{HTML}{0072B2}
\definecolor{cborange}{HTML}{E69F00}
\definecolor{cbgreen}{HTML}{009E73}
\definecolor{cbpink}{HTML}{CC79A7}
\definecolor{cbred}{HTML}{D55E00}
\definecolor{cbgray}{HTML}{999999}
\definecolor{trailred}{HTML}{C0392B}
\tikzset{
  bpbox/.style={draw=#1, fill=#1!8, rounded corners=3pt, line width=0.8pt, align=center, inner sep=4pt},
  bpbox/.default=darkgray,
  bplab/.style={font=\scriptsize\itshape, text=black!65, align=center},
  bpflow/.style={->, line width=0.9pt, black!70},
  bpimg/.style={inner sep=0pt, draw=#1, line width=1.4pt},
  bpimg/.default=black!40}
\tikzset{
  card/.style={draw=black!22, fill=white, rounded corners=5pt, line width=0.6pt,
               drop shadow={shadow xshift=0.6pt, shadow yshift=-0.8pt, opacity=0.18}},
  step/.style={circle, fill=#1, text=white, font=\scriptsize\bfseries, inner sep=1.3pt, minimum size=11pt},
  pill/.style={draw=#1, text=#1!75!black, fill=white, rounded corners=5pt, font=\scriptsize, inner sep=2pt,
               line width=0.6pt},
  arr/.style={-{Stealth[length=2.2mm, width=1.8mm]}, line width=0.8pt, black!55},
  tag/.style={font=\scriptsize, text=black!55, align=center}}

\newcommand{\asm}[1]{\hyperref[asm:#1]{#1}}

%% file: sec/0_abstract.tex
\begin{abstract}
Diagnosing whole-slide images (WSIs) requires hypothesis-driven reasoning: pathologists form a differential diagnosis, seek discriminative evidence, weigh its reliability, and stop once it suffices. However, existing training-free WSI agents lack an explicit framework for such reasoning. They typically use a large language model (LLM) to aggregate the region descriptions of a multimodal LLM (MLLM), without a belief over the candidates or regard for the descriptions' reliability. Consequently, a repeated model error counts as independent evidence for an incorrect diagnosis. We introduce \method, a training-free WSI agent that casts the pathologist's Scan--Look--Weigh--Decide process as reliability-aware Bayesian reasoning. An explicit belief over the candidates is updated with each observation through a reliability-aware likelihood and guides where to look next and when to stop. This likelihood corrects for the models' answer bias, discounts correlated observations, and captures diagnostic confusions estimated without labels. The updates form an interpretable evidence trail of how each observation shapes the belief. On five MultiPathQA tasks spanning slide classification and slide VQA, \method achieves the highest average accuracy among training-free agents, 58.6\% against 34.1\% for the next best, with large gains on slide classification. On identical observations, Bayesian updating outperforms LLM aggregation on average, even with a larger LLM. \method's belief is also better calibrated than naive Bayes, so uncertain cases can be deferred to pathologists.
Code is available at \mbox{\url{https://github.com/bryanwong17/BayesPath}}.
\end{abstract}

%% file: sec/1_intro.tex
\section{Introduction}
\label{sec:intro}

Whole-slide image (WSI) analysis is central to computational pathology, yet a single slide can span $100{,}000\times100{,}000$ pixels~\cite{slidechat,pathagent}, making exhaustive high-resolution inspection impractical. Diagnosis therefore requires locating sparse diagnostic regions and integrating evidence across magnifications~\cite{evipathbench}. Pathologists do this through a hypothesis-driven process: they first \emph{scan} the slide at low magnification to form a \emph{differential diagnosis} (a short list of candidates), then \emph{look} for findings that distinguish them, \emph{weigh} their reliability, and \emph{decide} when the evidence is sufficient~\cite{elstein,ledley,pauker}. Each finding updates the differential, which guides what to examine next and when to stop.

Recent WSI methods follow two main paradigms. \textbf{Direct prediction methods} answer from a fixed input: slide-level models~\cite{abmil,clam,transmil,titan,slidechat,wsillava} aggregate patch features into a slide representation, and multimodal large language models (MLLMs), such as Quilt-LLaVA~\cite{quiltllava} and Patho-R1~\cite{pathor1}, answer from a thumbnail or selected patches. \textbf{WSI agents} instead gather evidence by exploring the slide step by step. Some learn this exploration from curated navigation instructions or pathologists' viewing records~\cite{cpathagent,adaptivepath,pathpilot}, which are costly to collect. Training-free WSI agents, such as GIANT~\cite{giant}, PathAgent~\cite{pathagent} and PathNavigate~\cite{pathnavigate}, avoid this cost by building on frozen models. They typically select a series of regions, or \emph{views}, use a pathology MLLM to describe each view as an \emph{observation}, and rely on a large language model (LLM) to combine these observations into an answer, a step we term \emph{LLM aggregation}. Although this exploration resembles a pathologist's search, the reasoning behind it has two problems (\cref{fig:teaser}(a)).

\textbf{(i)~No explicit belief over candidate answers.} Without a distribution over the candidates, an agent lacks a principled basis for combining observations, selecting discriminative views and deciding when the evidence suffices. PathAgent thus leaves sufficiency to its LLM, which often stops prematurely: on 68\% of TCGA cancer-type questions, it stops after the first round, but only 25\% of these are correct (\cref{sec:confidence}).

\textbf{(ii)~No reliability-aware evidence integration.} Existing agents treat each observation as reliable, independent evidence. However, views from the same slide show the same tissue and are described by the same MLLM, so a systematic error can recur across views. In \cref{fig:teaser}(a), the MLLM repeatedly misreads ovarian carcinoma as lung adenocarcinoma, and the LLM counts each repeat as new evidence for the wrong diagnosis. Across TCGA, PathAgent and PathNavigate predict lung adenocarcinoma for 50\% and 54\% of slides, although only 5\% carry this label (\cref{sec:main}). A belief alone does not fix this: naive Bayes also counts each repeat as new evidence and grows confidently wrong (\cref{sec:reliab}).

\begin{figure*}[t]
\centering
\includegraphics[width=0.99\textwidth]{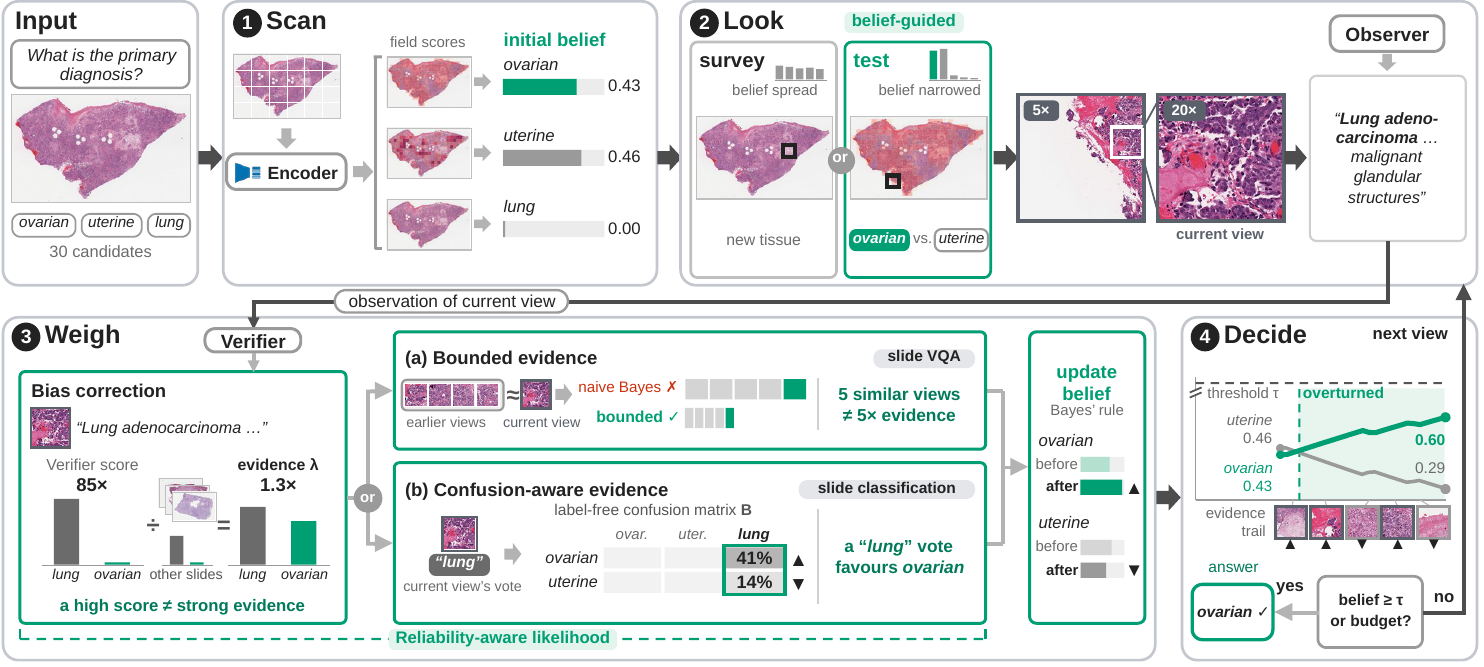}
\vspace{-0.5em}
\caption{\textbf{Overview of \method.} \textbf{(1)~Scan} forms the initial belief. \textbf{(2)~Look} surveys while the belief is spread, else tests the top candidates. \textbf{(3)~Weigh} corrects the Verifier's answer bias and uses (a)~bounded evidence for correlated views (VQA) or (b)~confusion-aware evidence for systematic confusions (classification). \textbf{(4)~Decide} answers once the top belief reaches $\tau$ or the budget ends, else reads the next view.}
\label{fig:method}
\vspace{-0.6em}
\end{figure*}

To address both limitations, we propose \method, a training-free WSI agent that casts the pathologist's \textbf{Scan--Look--Weigh--Decide} process as reliability-aware Bayesian reasoning (\cref{fig:teaser}(b), \cref{fig:method}). A pathology encoder first scores low-power fields to form an initial \emph{belief} over the candidates (Scan). This belief selects each next view, which a pathology MLLM, the \emph{Observer}, describes (Look). Each observation updates the belief by Bayes' rule through a \emph{reliability-aware likelihood} (Weigh), and the belief decides when to answer (Decide). The likelihood needs no labels; it only assumes unlabelled slides of the same task, abundant in hospital archives, to learn answer bias and confusions in classification (\cref{sec:impl}), while slide VQA uses bounded evidence from the slide alone. Our contributions are threefold:
\begin{itemize}[leftmargin=*,itemsep=1pt,topsep=2pt]
  \item We introduce \method, a training-free Bayesian WSI agent that keeps an explicit belief over the candidates to combine evidence and guide view selection and stopping, with an interpretable trail of how each view moves it.
  \item We develop a reliability-aware likelihood that corrects for model answer bias, discounts correlated observations through \emph{bounded evidence}, and models systematic confusions through label-free \emph{confusion-aware evidence}.
  \item Across five MultiPathQA tasks, \method achieves the highest average accuracy among training-free agents (58.6\% vs.\ 34.1\%), with large gains on slide classification. On identical observations, Bayesian updating outperforms LLM aggregation on average, even with a larger LLM. Its belief is also better calibrated than naive Bayes, and its lead holds with a stronger encoder.
\end{itemize}

%% file: sec/2_related.tex
\section{Related Work}
\label{sec:related}

\subsection{Multimodal Pathology Models}
Vision--language encoders such as PLIP~\cite{plip} and CONCH~\cite{conch} align pathology images with text. Pathology and biomedical MLLMs, including Quilt-LLaVA~\cite{quiltllava}, Patho-R1~\cite{pathor1} and LLaVA-Med~\cite{llavamed}, describe images and answer questions in free form, but typically see only a few patches or a low-resolution thumbnail. To cover the whole slide, SlideChat~\cite{slidechat} and WSI-LLaVA~\cite{wsillava} pass aggregated patch features to an LLM, and HistoSelect~\cite{actpath} first retrieves question-relevant patches with a trained selector. However, these models answer in a single pass, with no estimate of their reliability. \method instead uses a pathology encoder and MLLM as frozen evidence sources, choosing what they read and weighing each output by its reliability.

\subsection{Pathology Agents}
Pathology agents extend beyond single-pass analysis by iteratively examining slide regions. Some learn where to look: AdaptivePath~\cite{adaptivepath} and PathPilot~\cite{pathpilot} learn view selection from pathologist annotations or viewing behaviour, and other agents build on task-trained models~\cite{slideseek,cpathagent,pathfinder,mmnavagent}. Among training-free agents, GIANT~\cite{giant} lets a general-purpose MLLM navigate the slide, PathAgent~\cite{pathagent} retrieves question-relevant regions, describes them using a pathology MLLM and relies on an LLM to assess evidence sufficiency, and PathNavigate~\cite{pathnavigate} first scans for unusual regions before searching for question-relevant evidence. They aggregate observations with an LLM, without a belief over the candidates or a model of observation reliability. \method instead updates a Bayesian belief with a reliability-aware likelihood to guide view selection and stopping.

\subsection{Bayesian Reasoning and Source Reliability}
Medical diagnosis has long been viewed as a form of Bayesian reasoning. Clinicians refine differential diagnoses by evaluating competing hypotheses~\cite{elstein}, Bayes' rule formalises this process through probability updates~\cite{ledley}, and decision thresholds determine when the evidence is sufficient to act~\cite{pauker,wald1945}, as in selective prediction~\cite{chow,geifman}. Recent agents similarly maintain explicit beliefs to guide subsequent questions or examinations~\cite{medclarify,bedllm}, or to choose which regions to read before an LLM answers~\cite{wong2026evidence}; \method instead answers from a belief updated with every observation. LLMs also integrate sequential evidence poorly: they deviate from Bayesian posteriors and plateau early~\cite{qiu2025bayesian}.

However, Bayesian inference depends critically on the reliability of its likelihoods. Prior work has addressed three sources of unreliability separately. First, language models exhibit input-independent answer preferences that can be corrected without labelled data~\cite{zhao2021calibrate,zheng2024pride,saerens2002adjusting}. Second, treating correlated observations as independent leads to overcounting evidence, a problem that tempered Bayesian updates mitigate by reducing the influence of each observation~\cite{bissiri2016general}. Third, systematic diagnostic confusions can be estimated without ground-truth labels by exploiting agreement among multiple sources~\cite{dawidskene}. WSI agents face all three: their models favour some answers, their views come from the same slide, and their MLLM systematically confuses diagnoses. \method unifies these corrections in a reliability-aware likelihood.

%% file: sec/3_method.tex
\section{Method}
\label{sec:method}
\label{sec:bayes}
\label{sec:setup}
\method answers diagnostic questions about whole-slide images by maintaining a probability distribution over the candidate answers and updating it as evidence is gathered from the slide. Inspired by pathologists' diagnostic reasoning, it follows four stages: Scan, Look, Weigh and Decide (\cref{fig:method}; pseudocode in Supp.~Alg.~\ref{alg:bayespath}, notation in Supp.~\ref{sec:supp_math}).

\paragraph{Problem setup.} A question $x$ about a slide comes with $K$ candidate answers
$\mathcal{H}=\{h_1,\dots,h_K\}$, one of which is correct. In \emph{slide classification}, every slide of a task has the same candidates, such as the 30 cancer types of TCGA; in \emph{slide VQA}, each question has its own candidates. The slide is available as a thumbnail $v_0$, the 5$\times$ fields
$f\in\mathcal{F}$ of its tissue grid, and the $4\times4$ sub-fields of each field at 20$\times$
(Supp.~\ref{sec:supp_calibration}). \method reads the thumbnail and then up to $T$ further views $v_t$, each a field or a sub-field chosen by the \emph{Explorer}, a rule on the belief. It relies on three frozen models: a pathology vision--language \emph{encoder} that scores regions against the candidates, a pathology MLLM, the \emph{Observer}, that writes an observation $o_t$ for each view, and a text reranker, the \emph{Verifier}, that scores how well each observation supports each candidate.

\paragraph{The belief.} Instead of a text record of the observations, \method keeps the \emph{belief}
\begin{equation}
  b_t(k)=P(h_k\mid \text{scan},o_{0},\dots,o_{t}),
  \label{eq:beliefdef}
\end{equation}
the probability of each candidate given all evidence so far, which acts as the pathologist's differential diagnosis. \textbf{Scan} forms the belief (\cref{sec:scan}), \textbf{Look} uses it to choose the next view
(\cref{sec:look}), \textbf{Weigh} updates it after each observation (\cref{sec:likelihood}), and \textbf{Decide} uses
it to stop and answer (\cref{sec:conclude}).

\subsection{Scan: An Initial Belief at Low Power}
\label{sec:scan}
A pathologist starts by scanning the slide at low power. \method likewise uses the encoder, which scores every 5$\times$ field $f$ against the candidates with its logit scale $\kappa$~\cite{conch}:
\begin{equation}
  q^{\mathrm{enc}}(k\mid f)=\frac{\exp\big(\kappa\cos(e(f),e(h_k))\big)}{\sum_{j}\exp\big(\kappa\cos(e(f),e(h_j))\big)},
  \label{eq:enc}
\end{equation}
where $e(\cdot)$ denotes the image and text embeddings. As the encoder prefers some answers regardless of the image, we divide each score by $\bar q^{\,\mathrm{enc}}_k$, its average for candidate $k$ over the other half of the task's slides (cross-fitted, \cref{sec:impl}), and average over all fields (Supp.~\ref{sec:supp_calibration}). The initial belief is
\begin{equation}
  b_{-1}(k)\propto\lambda^{\mathrm{scan}}_k=\frac{1}{|\mathcal{F}|}\sum_{f\in\mathcal{F}}
  \frac{q^{\mathrm{enc}}(k\mid f)}{\bar q^{\,\mathrm{enc}}_k}.
  \label{eq:scan}
\end{equation}
The encoder and the Observer differ in model and input, so we treat them as conditionally independent, with prior $b_{-1}$ (Supp.~\ref{sec:supp_math}). In slide VQA, no other slide shares the candidates, so $b_{-1}$ is uniform.

\subsection{Look: Choosing and Describing the Next View}
\label{sec:look}
A pathologist first surveys the slide and then tests the competing diagnoses on the regions that separate them, zooming in
when needed. The Explorer has the same two modes, a \emph{survey} of new tissue and a \emph{test} of one candidate against its rivals, and chooses between them before each view $v_t$, $t\ge1$, from the current belief $b=b_{t-1}$.

\paragraph{Survey or test.} The Explorer surveys with probability equal to the normalised entropy of the belief,
\begin{equation}
  P(\text{survey})=\tilde H(b)=-\frac{1}{\log K}\sum_{k=1}^{K}b(k)\log b(k),
  \label{eq:survey}
\end{equation}
and tests otherwise: it mostly surveys while the diagnosis is open (high entropy) and mostly tests as the belief narrows (low entropy). A survey reads a random unread field, since an open belief is not yet a reliable guide to where to look, and extends coverage beyond the tissue favoured in Scan. A test draws a candidate $h_k$ from the belief, $k\sim b$, as in Thompson sampling~\cite{thompson}: rather than always testing the leading candidate, it tests each in proportion to its belief, so a wrong early lead can still be overturned.

\paragraph{Where to test.} A good view for testing $h_k$ is one where $h_k$ stands out from its plausible rivals,
not merely one that fits $h_k$. The set $\mathcal{D}$ holds $h_k$ and its rivals, the fewest top candidates (at least two) whose belief sums to at
least $m$. The Explorer scores each field by the encoder's probability of $h_k$ within $\mathcal{D}$ and
takes the best:
\begin{equation}
  \begin{gathered}
  \rho_k(v)=\frac{\exp\big(\kappa\cos(e(v),e(h_k))\big)}{\sum_{j\in\mathcal{D}}\exp\big(\kappa\cos(e(v),e(h_j))\big)},\\
  f^\star=\argmax_{f\in\mathcal{O}}\rho_k(f),
  \end{gathered}
  \label{eq:look}
\end{equation}
where the \emph{open} fields $\mathcal{O}$ are those with a view left to read: if $f^\star$ is unread, the Explorer reads it at 5$\times$; otherwise, it zooms in to its unread 20$\times$ sub-field with the highest $\rho_k$.

\paragraph{Describe.} The Observer receives the view, its coordinates and magnification, the question and the candidates,
and writes the observation $o_t$.

\subsection{Weigh: A Reliability-Aware Likelihood}
\label{sec:weigh}
\label{sec:likelihood}
If every observation counts as independent evidence, the models' answer bias, the correlated views of one slide and the Observer's systematic errors make the belief confident in wrong answers (\cref{sec:reliab}). \method therefore applies \textbf{bias correction}, then one of two forms of evidence: \textbf{(a)~bounded evidence} $L^{\mathrm{bound}}$ (\cref{eq:update}), which caps the evidence of correlated views (slide VQA), or \textbf{(b)~confusion-aware evidence} $L^{\mathrm{conf}}$ (\cref{eq:confusion}), which is also bounded and models the Observer's systematic confusions, estimated from other slides with the same candidates (slide classification).

\paragraph{Bias correction: from a score to evidence.} Every model has an \emph{answer bias}: it
prefers some answers regardless of the input. If a model names lung adenocarcinoma on half of all slides, whatever their diagnosis, naming it again is weak evidence for this slide. We read the score as a posterior under the model's implicit prior $\bar q$
(Supp.~\ref{sec:supp_math}). By Bayes' rule, dividing by this prior gives a likelihood
ratio~\cite{bourlard},
\begin{equation}
  \lambda_k(o)=\frac{q(k\mid o)}{\bar q_k},
  \label{eq:inversion}
\end{equation}
so $\lambda_k(o)>1$ means $o$ supports $h_k$ more than a typical observation does. We estimate $\bar q$ without
labels as the model's average score on the \emph{other} slides with the same candidates
(cross-fitted, \cref{sec:impl}), the standard label-free \emph{bias correction}~\cite{saerens2002adjusting,zhao2021calibrate}. In slide VQA, no other slide shares the candidates, so $\bar q$ is uniform.

\paragraph{Scoring an observation.} An observation is free text that may name one candidate, several or
none, and the features it describes can support a candidate it does not name. The Verifier, a text reranker~\cite{qwen3reranker}, therefore scores it against every candidate. For
each candidate $h_k$, it reads the question, the observation and $h_k$, and returns the probability of ``yes'', $p_{\mathrm{yes}}(o_t,h_k)$ (Supp.~\ref{sec:supp_prompts}). We normalise over the candidates and remove the
Verifier's answer bias with \cref{eq:inversion}:
\begin{equation}
  q(k\mid o_t)=\frac{p_{\mathrm{yes}}(o_t,h_k)}{\sum_{j=1}^{K}p_{\mathrm{yes}}(o_t,h_j)},\qquad
  \lambda_{tk}=\frac{q(k\mid o_t)}{\bar q_k}.
  \label{eq:verifier}
\end{equation}

\paragraph{(a) Bounded evidence for correlated views.} The views of one slide show the same tissue and are described by the same Observer, so they are correlated~\cite{transmil}. Multiplying their ratios, as naive Bayes does ($L^{\mathrm{NB}}_{tk}=\prod_{s\le t}\lambda_{sk}$), counts one finding many times, and a repeated mistake makes the belief confidently wrong (\cref{sec:reliab}). As in multiple-instance learning (MIL) for WSIs~\cite{abmil,clam,hivemil}, we treat the slide as a bag of views: each of its $T{+}1$ views has prior probability $1/(T{+}1)$ of carrying the slide's evidence, and unread views carry none ($\lambda=1$), our multiple-instance assumption (Supp.~\ref{sec:supp_math}). Averaging over the views gives
\begin{equation}
  L^{\mathrm{bound}}_{tk}=1+\frac{1}{T{+}1}\sum_{s=0}^{t}\big(\lambda_{sk}-1\big)
  \label{eq:update}
\end{equation}
(derivation in Supp.~\ref{sec:supp_update}). It is \textbf{permutation-invariant}, as the order of the views does not matter, and \textbf{bounded}: no set of views supports $h_k$ more than its most supportive view, so similar views count at most once.

\paragraph{(b) Confusion-aware evidence for systematic confusions.} Bias correction discounts answers named on every slide, but the Observer also confuses specific diagnoses with each other. To count these confusions, \method reduces each view to a \emph{vote}, the candidate it supports most, $c_t=\argmax_k\lambda_{tk}$, and describes the votes by a confusion
matrix $B$, where $B_{kj}=P(c=j\mid h_k)$ is the probability that a view of a slide with answer $h_k$ votes for $h_j$.

\paragraph{Estimating the confusions without labels.} Crowdsourcing estimates the error rates of annotators from how they agree~\cite{dawidskene}. We estimate $B$ in the same way, with the expectation--maximisation (EM) algorithm~\cite{dempster1977em} over the unlabelled slides $i$ of the task that share the candidates (cross-fitted, \cref{sec:impl}). Let $F_{ij}$ be the
fraction of the views of slide $i$ that vote $j$, and $W_{ik}$ the estimated probability that slide $i$ has answer $h_k$, which is never observed.
Starting from the initial belief, $W_{ik}=b^{(i)}_{-1}(k)$, EM alternates
\begin{equation}
  \begin{aligned}
  \text{E:}\quad W_{ik}&\propto b^{(i)}_{-1}(k)\exp\Big(\textstyle\sum_{j}F_{ij}\log B_{kj}\Big),\\
  \text{M:}\quad B_{kj}&\propto\tfrac{1}{K}+\textstyle\sum_i W_{ik}F_{ij}.
  \end{aligned}
  \label{eq:em}
\end{equation}
The E-step estimates each slide's answer from its votes and the current confusions; the M-step re-estimates each row of $B$ from the votes weighted by these estimates, plus one pseudo-count spread over each row (derivation in Supp.~\ref{sec:supp_habits}).

\paragraph{Using the confusions.} $B$ gives the likelihood of each vote under each answer: a view of an $h_k$ slide votes $c$ with probability $B_{kc}$. A vote is thus evidence for every answer under which it is likely, not only for the answer it names. For example, views of ovarian slides vote ``lung'' 41\% of the time and views of uterine slides 14\% (\cref{fig:method}(b)), so a ``lung'' vote favours ovarian. A vote equally likely under all answers carries no evidence. As in (a), the votes of one slide are correlated and together should count as only one observation, so each vote's likelihood enters with the power $1/(T{+}1)$:
\begin{equation}
  L^{\mathrm{conf}}_{tk}=\prod_{s=0}^{t}B_{k c_s}^{1/(T+1)}.
  \label{eq:confusion}
\end{equation}
Like \cref{eq:update}, this evidence is permutation-invariant and bounded, and unread views add nothing (Supp.~\ref{sec:supp_update}). It is a tempered Bayesian update~\cite{bissiri2016general} whose exponent is set by the view budget, so no weight is tuned.

\paragraph{Sequential Bayesian update.} Bayes' rule combines the initial belief with the evidence $L_{tk}$ of the views so far (\cref{eq:update} or \cref{eq:confusion}): $b_t(k)\propto b_{-1}(k)\,L_{tk}$. Look and Decide use the belief after every view, so we update it from the previous belief, $b_{t-1}(k)\propto b_{-1}(k)\,L_{t-1,k}$. Their ratio cancels $b_{-1}$:
\begin{equation}
  b_t(k)\ \propto\ b_{t-1}(k)\,\frac{L_{tk}}{L_{t-1,k}},\qquad L_{-1,k}=1.
  \label{eq:belief}
\end{equation}
Each new view thus weighs every candidate $h_k$ by the evidence it adds, $L_{tk}/L_{t-1,k}$: the candidates it supports more than their rivals gain belief, and the others lose it.
In slide classification, where $B$ is estimated from the other slides, $L_{tk}=L^{\mathrm{conf}}_{tk}$; in slide VQA, $L_{tk}=L^{\mathrm{bound}}_{tk}$, and \cref{eq:belief} is exact Bayes' rule under the multiple-instance assumption (derivation in Supp.~\ref{sec:supp_update}).

\subsection{Decide: When to Answer}
\label{sec:conclude}
A pathologist commits to a diagnosis once it is clear. Likewise, \method checks the belief after Scan and after every view. As in sequential hypothesis testing~\cite{wald1945}, it stops at the first step $t^\star$ at which the top candidate reaches belief $\tau$, or after $T$ views, and answers the most probable candidate:
\begin{equation}
  \begin{aligned}
  t^\star&=\min\bigl\{t\ge -1:\ \max_k b_t(k)\ge\tau\ \text{or}\ t=T\bigr\},\\
  \hat y&=\argmax_k\, b_{t^\star}(k).
  \end{aligned}
  \label{eq:decide}
\end{equation}
Each answer comes with an \emph{evidence trail}: every view read, its magnification and observation,
and how it changed the belief, which shows the views that decided the answer.

%% file: sec/4_experiments.tex
\section{Experiments}
\label{sec:exp}
\begin{table*}[t]
\centering
\footnotesize
\setlength{\tabcolsep}{4.5pt}
\renewcommand{\arraystretch}{0.84}
\caption{\textbf{Zero-shot comparison} on the five MultiPathQA tasks (\%). BACC: balanced accuracy (mean per-class accuracy); ACC: accuracy. $K$: number of candidates. Best in \textbf{bold}, second \underline{underlined}; $^*$trained on in-domain (TCGA) data, for reference. Mean $\pm$ std of three runs.}
\label{tab:main}
\vspace{-0.6em}
\renewcommand{\pms}[2]{#1{\scriptsize\color{gray!85!black}$\pm$#2}}
\newcommand{\K}[1]{{\scriptsize\color{gray!85!black}$K{=}#1$}}
\newcommand{\ours}[1]{\cellcolor{cbblue!8}#1}
\newcommand{\V}[1]{{\scriptsize #1}}
\newcommand{\tbd}{{\color{gray!70}--}}
\begin{tabular}{@{}lllcccccc}
\toprule
 & & & \multicolumn{3}{c}{Slide classification (BACC)} & \multicolumn{2}{c}{Slide VQA (ACC)} & \\
\cmidrule(lr){4-6}\cmidrule(lr){7-8}
Category & Method & Backbone & TCGA & GTEx & PANDA & ExpertVQA & SlideBench & Avg. \\
 & & & \K{30} & \K{20} & \K{6} & \K{4} & \K{4} & \\
\midrule
\multirow{3}{*}{\begin{tabular}[c]{@{}l@{}}\textit{General MLLM}\\{\scriptsize\color{gray!85!black}Input: thumbnail}\end{tabular}} & Pixtral-12B~\cite{pixtral} & \V{Mistral-NeMo-12B} & \pms{5.1}{1.2} & \pms{11.0}{3.7} & \pms{15.0}{2.3} & \pms{35.2}{2.1} & \pms{45.5}{1.9} & \pms{22.4}{2.0} \\
 & Gemma-3-27B~\cite{gemma3} & \V{--} & \pms{7.6}{0.4} & \pms{16.8}{1.8} & \pms{17.5}{0.8} & \pms{37.2}{2.5} & \pms{30.5}{3.2} & \pms{21.9}{0.0} \\
 & Qwen3.8-27B~\cite{qwen38} & \V{--} & \pms{8.3}{3.6} & \pms{29.0}{4.5} & \pms{18.2}{1.7} & \pms{41.7}{1.6} & \pms{45.5}{0.6} & \pms{28.5}{0.5} \\
\midrule
\multirow{4}{*}{\begin{tabular}[c]{@{}l@{}}\textit{Medical MLLM}\\{\scriptsize\color{gray!85!black}Input: thumbnail}\end{tabular}} & LLaVA-Med-7B~\cite{llavamed} & \V{Mistral-7B} & \pms{2.9}{0.4} & \pms{6.9}{1.9} & \pms{10.2}{1.3} & \pms{29.7}{1.4} & \pms{31.8}{0.3} & \pms{16.3}{0.6} \\
 & Quilt-LLaVA-7B~\cite{quiltllava} & \V{LLaVA-1.5-7B} & \pms{3.1}{1.3} & \pms{9.3}{0.4} & \pms{14.2}{1.9} & \pms{43.0}{1.6} & \pms{42.0}{3.1} & \pms{22.3}{1.6} \\
 & MedGemma-27B~\cite{medgemma} & \V{Gemma-3-27B} & \pms{5.1}{1.5} & \pms{7.9}{0.6} & \pms{15.5}{3.5} & \pms{35.2}{3.6} & \pms{45.9}{0.3} & \pms{21.9}{1.5} \\
 & Patho-R1-7B~\cite{pathor1} & \V{Qwen2.5-VL-7B} & \pms{4.0}{0.2} & \pms{36.2}{2.4} & \pms{15.1}{6.4} & \pms{39.8}{5.5} & \pms{41.3}{1.3} & \pms{27.3}{2.4} \\
\midrule
\multirow{3}{*}{\begin{tabular}[c]{@{}l@{}}\textit{Supervised}\\\textit{WSI model}\\{\scriptsize\color{gray!85!black}Input: slide}\end{tabular}} & WSI-VQA~\cite{wsivqa} & \V{ResNet-50} & \pms{3.9}{0.0} & \pms{5.0}{0.0} & \underline{\smash{\pms{21.4}{0.0}}} & \pms{25.0}{0.0} & \pms{18.8}{0.0} & \pms{14.8}{0.0} \\
 & SlideChat~\cite{slidechat} & \V{CONCH\,+\,Qwen2.5-7B} & \pms{1.9}{0.4} & \pms{5.1}{0.2} & \pms{17.1}{2.2} & \pms{35.4}{4.0} & \pms{66.3}{2.1}$^*$ & \pms{25.2}{1.0} \\
 & WSI-LLaVA~\cite{wsillava} & \V{GigaPath\,+\,Vicuna-7B} & \pms{10.1}{0.8} & \pms{16.4}{0.8} & \pms{15.3}{0.9} & \pms{38.8}{1.2} & \pms{60.1}{2.6}$^*$ & \pms{28.1}{0.5} \\
\midrule
\multirow{4}{*}{\begin{tabular}[c]{@{}l@{}}\textit{Training-free}\\\textit{WSI agent}\\{\scriptsize\color{gray!85!black}Input: slide}\end{tabular}} & GIANT~\cite{giant} & \V{Patho-R1-7B} & \pms{3.7}{0.4} & \pms{10.0}{2.5} & \pms{16.3}{1.6} & \pms{37.8}{6.6} & \pms{51.3}{3.0} & \pms{23.8}{2.1} \\
 & PathAgent~\cite{pathagent} & \V{CONCH\,+\,Patho-R1-7B\,+\,Qwen3-4B} & \underline{\smash{\pms{14.5}{1.0}}} & \underline{\smash{\pms{45.0}{3.3}}} & \pms{20.2}{0.9} & \pms{38.5}{3.2} & \pms{52.3}{0.9} & \underline{\smash{\pms{34.1}{1.5}}} \\
 & PathNavigate~\cite{pathnavigate} & \V{CONCH\,+\,Patho-R1-7B\,+\,Qwen3-4B} & \pms{6.2}{1.1} & \pms{33.3}{2.8} & \pms{17.2}{1.1} & \underline{\smash{\pms{46.9}{0.8}}} & \underline{\smash{\pms{57.5}{0.6}}} & \pms{32.2}{0.4} \\
\cmidrule{2-9}
 & \ours{\textbf{\method (ours)}} & \ours{\V{CONCH\,+\,Patho-R1-7B\,+\,Qwen3-4B}} & \ours{\textbf{\pms{75.3}{0.7}}} & \ours{\textbf{\pms{83.4}{0.9}}} & \ours{\textbf{\pms{28.2}{0.2}}} & \ours{\textbf{\pms{47.7}{4.3}}} & \ours{\textbf{\pms{58.4}{0.9}}} & \ours{\textbf{\pms{58.6}{0.9}}} \\
\bottomrule
\end{tabular}
\vspace{-0.6em}
\end{table*}

\subsection{Experimental Setup}
\label{sec:impl}
\label{sec:data}
\paragraph{Datasets and metrics.} We evaluate \method on five public MultiPathQA tasks~\cite{giant} (details in Supp.~\ref{sec:supp_tasks}). Three involve slide classification: \textbf{TCGA}~\cite{tcga}, classifying 30 cancer types; \textbf{GTEx}~\cite{gtex}, identifying 20 organs from normal tissue; and \textbf{PANDA}~\cite{panda}, grading prostate cancer as benign or ISUP grades 1--5. The remaining two are four-choice slide VQA tasks: \textbf{ExpertVQA}~\cite{giant}, with pathologist-written questions on TCGA slides, and \textbf{SlideBench}, from SlideBench-VQA (TCGA)~\cite{slidechat}. We report balanced accuracy (BACC) for classification and accuracy (ACC) for slide VQA, averaged over three seeds, and assess calibration and selective prediction with expected calibration error (ECE) and area under the risk--coverage curve (AURC)~\cite{geifman}; lower is better.

\paragraph{Baselines.} We compare \method with four groups of methods: general MLLMs and medical MLLMs (both answering from the slide thumbnail), supervised WSI models, and the training-free WSI agents GIANT~\cite{giant}, PathAgent~\cite{pathagent} and PathNavigate~\cite{pathnavigate} (details in Supp.~\ref{sec:supp_models}).

\paragraph{Implementation.} To ensure a fair comparison, all agents use frozen models: \method uses CONCH~\cite{conch} as encoder, Patho-R1-7B~\cite{pathor1} as Observer and Qwen3-Reranker-4B~\cite{qwen3reranker} as Verifier; PathAgent and PathNavigate use the same encoder and Observer, with the LLM Qwen3-4B (same size and family as the Verifier) to combine observations; GIANT uses Patho-R1-7B. \cref{sec:titan_main} replaces CONCH with the stronger CONCH-Z~\cite{conchz} in the three agents that use an encoder. We set $T=25$, $m=0.95$ and $\tau=0.95$ for all five tasks (analysis of $\tau$ in Supp.~\ref{sec:supp_anytime}). The answer bias $\bar q$ and confusion matrix $B$ are estimated from unlabelled slides by cross-fitting: each half of a task's slides is evaluated with statistics from the other half, so no slide contributes to its own prediction (Supp.~\ref{sec:supp_impl}).

\begin{table}[t]
\centering
\footnotesize
\setlength{\tabcolsep}{2.6pt}
\caption{\textbf{Ablation of \method} (\%). Combine: rule that combines the Scan and observations. BC: bias correction (with BC alone, Bayes' rule is naive Bayes). BE: bounded evidence. CE: confusion-aware evidence. BV: belief-guided views (else question-similarity views). $^\dagger$Same as \method, which uses BE on slide VQA.}
\label{tab:memory}
\label{tab:look}
\label{tab:weigh}
\vspace{-0.6em}
\resizebox{\linewidth}{!}{%
\begin{tabular}{@{}lcccccccccc@{}}
\toprule
 & & & & & \multicolumn{3}{c}{Slide classification} & \multicolumn{2}{c}{Slide VQA} & \\
\cmidrule(lr){6-8}\cmidrule(lr){9-10}
Combine & BC & BE & CE & BV & TCGA & GTEx & PANDA & \shortstack{Expert\\VQA} & \shortstack{Slide\\Bench} & Avg. \\
\midrule
Qwen3-4B & & & & \ding{51} & 24.8 & 55.1 & 21.5 & 43.8 & 55.7 & 40.2 \\
Qwen3.8-27B & & & & \ding{51} & 36.8 & 68.0 & 22.1 & \textbf{48.7} & 55.8 & 46.3 \\
\midrule
Bayes' rule & \ding{51} & & & \ding{51} & 36.0 & 65.9 & 21.9 & 44.3 & 54.7 & 44.6 \\
Bayes' rule & \ding{51} & \ding{51} & & \ding{51} & 73.0 & 79.8 & 26.2 & \textcolor{cbgray}{47.7}$^\dagger$ & \textcolor{cbgray}{58.4}$^\dagger$ & 57.0 \\
Bayes' rule & \ding{51} & \ding{51} & \ding{51} & & 74.2 & 83.1 & 27.3 & 43.5 & 57.9 & 57.2 \\
\rowcolor{cbblue!7}\method & \ding{51} & \ding{51} & \ding{51} & \ding{51} & \textbf{75.3} & \textbf{83.4} & \textbf{28.2} & 47.7 & \textbf{58.4} & \textbf{58.6} \\
\bottomrule
\end{tabular}}
\vspace{-0.8em}
\end{table}

\begin{figure*}[t]
\centering
\includegraphics[width=\textwidth]{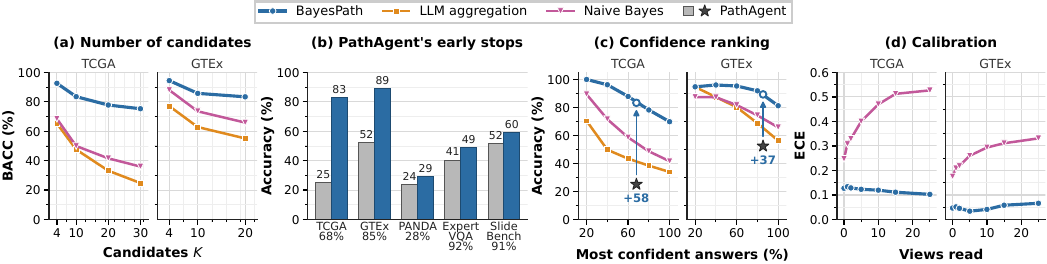}
\vspace{-1.8em}
\caption{\textbf{\method versus LLM reasoning and naive Bayes.} (a)~\method leads for every number of candidates. (b)~PathAgent stops too early: its first-round answers (grey) are less accurate than \method's most confident ones on as many questions (blue). (c)~Ranked by confidence, \method's top answers are the most accurate. (d)~Naive Bayes grows overconfident; \method stays calibrated.}
\label{fig:belief}
\label{fig:decide}
\label{fig:accum}
\vspace{-0.8em}
\end{figure*}

\subsection{Comparison with Existing Methods}
\label{sec:main}
\paragraph{Overall results.} As shown in \cref{tab:main}, \method achieves the highest average accuracy of 58.6\%, outperforming the strongest competing training-free agent, PathAgent, by 24.5 points. The best MLLM on average, Qwen3.8-27B, trails by 30.1 points, as thumbnails lose the cellular detail needed to tell the candidates apart. PathAgent and PathNavigate use the same encoder and Observer as \method, so the gains come from how \method reasons: mainly from reliability-aware Bayesian aggregation (\cref{sec:bayesexp,sec:reliab}), with belief-guided views adding gains on all five tasks (\cref{tab:memory}).

\paragraph{The gain is largest on slide classification.} On TCGA, \method achieves 75.3\% balanced accuracy across 30 cancer types, compared with 14.5\% for the strongest competing agent, PathAgent. On GTEx, it reaches 83.4\% across 20 organs, compared with 45.0\% (McNemar $p<10^{-10}$ against both agents; Supp.~\ref{sec:supp_stats}). In slide classification, all slides share the same question, so systematic Observer errors recur across views and slides. Existing agents count each recurrence as new evidence, so one misreading dominates: PathAgent and PathNavigate predict lung adenocarcinoma for 50\% and 54\% of TCGA slides, although only 5\% belong to this class (Supp.~\ref{sec:supp_spread}). \method instead discounts these recurrences and models the confusions (\cref{sec:reliab}), so it matches or outperforms both agents on 29 of 30 cancer types and 18 of 20 organs (Supp.~\cref{fig:percls}). \method also leads the strongest competing agent by 8.0 points on PANDA's six grades and achieves the highest zero-shot accuracy on both slide VQA tasks, where only bounded evidence applies.

\subsection{The Role of the Belief}
\label{sec:bayesexp}
\label{sec:memory}
\label{sec:confidence}

\paragraph{The belief combines the evidence.} Existing agents combine observations with an LLM, \method with Bayes' rule. To compare the two rules on the same evidence, an LLM answers from \method's own Scan and observations (\cref{tab:memory}). With Qwen3-4B, as large as the Verifier, \method leads on all five tasks by 18.4 points on average, from 2.7 on SlideBench to 50.5 on TCGA, where the same misreadings recur across views and slides. Within a slide, the LLM takes repeated misreadings for agreement; reading one slide at a time, it cannot see an answer bias across slides. \method corrects both (\cref{sec:reliab}). A ${\sim}7\times$ larger LLM does not close the gap: Qwen3.8-27B still trails by 12.3 points on average, and by 10.4 when prompted to keep a belief and discount repeats (Supp.~\ref{sec:supp_prompted}). Fewer candidates do not close it either: with four candidates, \method still leads Qwen3-4B by 27.3 points on TCGA and 17.3 on GTEx, and its lead grows with more candidates (\cref{fig:belief}(a); Supp.~\ref{sec:supp_fewer}). Unlike an LLM, Bayes' rule is order-invariant: reversing the views changes 21--33\% of Qwen3-4B's answers and 7--26\% of Qwen3.8-27B's, but none of \method's (Supp.~\ref{sec:supp_order}).

\paragraph{The belief chooses where to look.} Besides combining evidence, the belief picks the next region to read. Replacing these belief-guided views with question-similarity views, as in existing agents, reduces accuracy on all five tasks, by 1.4 points on average and 4.2 on ExpertVQA (\cref{tab:memory}). At 20$\times$, belief-guided views point to the true answer more often than question-similarity views: 40.4\% vs.\ 23.8\% on TCGA, 47.5\% vs.\ 39.6\% on GTEx, 35.1\% vs.\ 32.2\% on ExpertVQA and 59.9\% vs.\ 50.5\% on SlideBench (Supp.~\ref{sec:supp_zoom}).

\paragraph{The belief knows when the evidence suffices.} PathAgent has no belief: after each round, its LLM judges whether to stop and can take a repeated error for consistent evidence, so it stops too early. On TCGA, it stops after the first round on 68\% of questions, and these are 25.1\% correct, barely above its 23.1\% overall accuracy. On the 68\% of questions where \method is most confident, its answers are 83.3\% correct, and it leads this comparison on all five tasks (\cref{fig:belief}(b)). It also shows which answers to trust: on TCGA and GTEx, \method's most confident answers are the most accurate (\cref{fig:belief}(c)), and on TCGA its top 40\% are 96.2\% correct, vs.\ 71.6\% and 49.8\% for naive Bayes and LLM aggregation.

\subsection{Reliability-Aware Likelihood}
\label{sec:reliab}

\begin{table}[t]
\centering
\footnotesize
\setlength{\tabcolsep}{3pt}
\caption{\textbf{Trusting the belief} on \method's evidence; $\downarrow$: lower is better. LLM aggregation keeps no belief, so it has no ECE.}
\label{tab:trust}
\vspace{-0.6em}
\resizebox{\linewidth}{!}{%
\begin{tabular}{@{}llccccc@{}}
\toprule
 & & \multicolumn{3}{c}{Slide classification} & \multicolumn{2}{c}{Slide VQA} \\
\cmidrule(lr){3-5}\cmidrule(l){6-7}
Metric & Rule & TCGA & GTEx & PANDA & ExpertVQA & SlideBench \\
\midrule
\multirow{2}{*}{ECE $\downarrow$} & Naive Bayes & 0.527 & 0.331 & 0.671 & 0.545 & 0.436 \\
 & \cellcolor{cbblue!7}\method & \cellcolor{cbblue!7}\textbf{0.103} & \cellcolor{cbblue!7}\textbf{0.066} & \cellcolor{cbblue!7}\textbf{0.127} & \cellcolor{cbblue!7}\textbf{0.116} & \cellcolor{cbblue!7}\textbf{0.088} \\
\midrule
\multirow{3}{*}{AURC $\downarrow$} & LLM aggregation & 0.488 & 0.179 & 0.756 & 0.399 & 0.268 \\
 & Naive Bayes & 0.329 & 0.186 & 0.772 & 0.463 & 0.367 \\
 & \cellcolor{cbblue!7}\method & \cellcolor{cbblue!7}\textbf{0.110} & \cellcolor{cbblue!7}\textbf{0.084} & \cellcolor{cbblue!7}\textbf{0.728} & \cellcolor{cbblue!7}\textbf{0.376} & \cellcolor{cbblue!7}\textbf{0.267} \\
\bottomrule
\end{tabular}}
\vspace{-0.8em}
\end{table}
\begin{figure*}[t]
\centering
\includegraphics[width=0.99\textwidth]{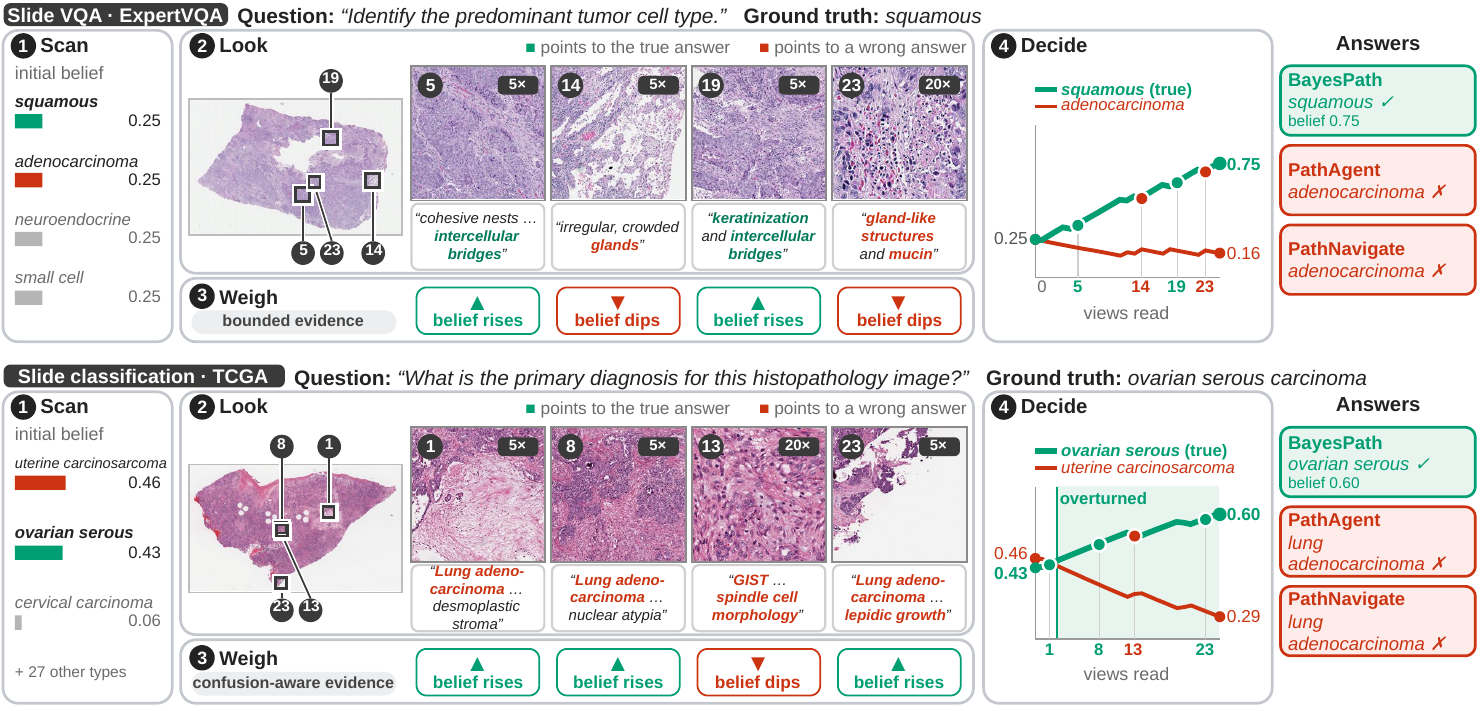}
\vspace{-0.5em}
\caption{\textbf{Evidence trails of \method.} PathAgent and PathNavigate trust every MLLM reading and answer with its misreading. \textbf{Top, bounded evidence}: views misread as \textcolor[HTML]{CC3311}{adenocarcinoma} barely lower \textcolor[HTML]{007A5A}{squamous}, while 20 squamous views raise it (0.25$\to$0.75). \textbf{Bottom, confusion-aware evidence}: a label-free confusion lets views misread as \textcolor[HTML]{CC3311}{lung adenocarcinoma} raise \textcolor[HTML]{007A5A}{ovarian serous} (0.43$\to$0.60).}
\label{fig:steps}
\vspace{-0.8em}
\end{figure*}
\paragraph{A belief is only as reliable as its evidence.} Even naive Bayes, the simplest belief, beats Qwen3-4B aggregation on average (\cref{tab:memory}). But a belief inherits its views' errors, which take three forms. First, the Verifier favours some answers regardless of the image. Second, the Observer repeats misreadings: on 80--95\% of slides per task, at least three views favour the same wrong answer, and on TCGA, 44\% of views favour lung adenocarcinoma, the label of only 5\% of slides (Supp.~\ref{sec:supp_repeat}). Third, some misreadings are systematic, \eg ovarian read as lung. Naive Bayes fixes only the first.

\vspace{-0.45em}
\paragraph{Each part of the likelihood corrects one error.} Bias correction removes the Verifier's answer bias, raising the raw product of its scores by 15.1 points on TCGA and 11.3 on GTEx (Supp.~\ref{sec:supp_parts}). Bounded evidence (\cref{eq:update}) caps what repeated misreadings can add and gives the largest gain over naive Bayes, from 3.4 points on ExpertVQA to 37.0 on TCGA (\cref{tab:memory}). The confusion matrix turns systematic misreadings into evidence, adding 2.0--3.6 points on classification; where no view names the truth, it raises accuracy from 23\% to 39\%, 17\% to 45\% and 14\% to 32\% on TCGA, GTEx and PANDA (Supp.~\ref{sec:supp_em}). Though label-free, the matrix correlates with the true confusions of held-out slides (Pearson $r=0.78$ on TCGA, $0.75$ on GTEx).

\vspace{-0.45em}
\paragraph{Bounded evidence also makes the confidence trustworthy.} Naive Bayes multiplies the views' likelihood ratios, so each repeat of an error counts again: its mean confidence reaches 89--99\%, and its ECE grows with more views on TCGA and GTEx (\cref{fig:belief}(d)). Sixteen copies of one view change 31--59\% of its classification answers and 9--27\% of Qwen3-4B's, but only 6--9\% of \method's (Supp.~\ref{sec:supp_copies}). The bound reduces this overconfidence without making \method underconfident: ECE, which penalises both, drops from 0.33--0.67 to 0.07--0.13 (Supp.~\ref{sec:supp_calib}). Its confidence also ranks answers better: AURC, which rewards ranking correct answers first, falls by 0.044--0.219 against naive Bayes (\cref{tab:trust}). On ExpertVQA and SlideBench, AURC falls by 0.087 and 0.100 (significant by bootstrap; Supp.~\ref{sec:supp_stats}).

\subsection{Integrating Pathology Foundation Models}
\label{sec:titan_main}
\begin{table}[t]
\centering
\footnotesize
\setlength{\tabcolsep}{3.5pt}
\renewcommand{\arraystretch}{0.92}
\caption{\textbf{A stronger encoder:} CONCH-Z in all three agents. Mean of three seeds (\%); standard deviations in Supp.~\cref{tab:supp_models}.}
\label{tab:stronger}
\vspace{-0.6em}
\resizebox{\linewidth}{!}{%
\begin{tabular}{@{}lccccc@{}}
\toprule
Method & TCGA & GTEx & PANDA & ExpertVQA & SlideBench \\
\midrule
PathAgent & 15.7 & 43.4 & 22.1 & 41.9 & 54.5 \\
PathNavigate & 6.3 & 30.8 & 18.8 & 47.1 & 58.0 \\
\rowcolor{cbblue!7}\method & \textbf{78.7} & \textbf{86.4} & \textbf{25.7} & \textbf{48.2} & \textbf{58.9} \\
\bottomrule
\end{tabular}}
\vspace{-1.1em}
\end{table}
\method's gains come from how it reasons, so they hold when its models are swapped without training. When PathAgent, PathNavigate and \method use the stronger CONCH-Z~\cite{conchz} as encoder, \method still leads on average, by wide margins on classification (\cref{tab:stronger}). The same holds with two other Observers, Quilt-LLaVA~\cite{quiltllava} and PathGen-LLaVA~\cite{pathgen}: \method reaches at least 71.5\% on TCGA, both agents at most 13.0\% (Supp.~\ref{sec:supp_swap}).

\subsection{Qualitative Evaluation}
\label{sec:case}
In both cases of \cref{fig:steps}, PathAgent and PathNavigate answer with the MLLM's misreading, whereas \method weighs each view by its reliability (more in Supp.~\ref{sec:supp_trails}). (a)~\emph{Bounded evidence}: four views misread as adenocarcinoma barely lower squamous, while 20 squamous views raise it from 0.25 to 0.75. (b)~\emph{Confusion-aware evidence}: the Scan favours uterine carcinosarcoma over the true ovarian serous, and three views are read as lung adenocarcinoma. Since \method learned that ovarian serous is often misread this way, these views raise it above the Scan's choice at view~2, to 0.60.

%% file: sec/5_conclusion.tex
\section{Conclusion}
\label{sec:conclusion}
We introduced \method, a training-free WSI agent that casts the pathologist's Scan--Look--Weigh--Decide process as reliability-aware Bayesian reasoning. Its explicit belief, updated by a likelihood learned from unlabelled slides, combines the evidence and guides view selection and stopping. On five tasks, \method has the highest average accuracy among training-free agents, beats LLM aggregation on average and is better calibrated. These findings show the value of explicit beliefs and reliability-aware likelihoods for accurate, calibrated and interpretable diagnosis.\label{endmain}

%% file: sec/6_llm_statement.tex
\section*{LLM Usage Statement}
LLMs and MLLMs are components of the method and baselines studied in this paper, as described in \cref{sec:method,sec:exp}. In addition, an LLM-based coding assistant (Claude Code, Anthropic) was used to assist with literature search, analysis, mathematical derivations, creating the figures, and drafting and editing the manuscript, including grammar. The authors designed the research, verified all results, and take full responsibility for the content of the paper.

%% file: sec/X_suppl.tex
\clearpage
\maketitlesupplementary
\appendix
\makeatletter\setlength{\@fptop}{0pt}\setlength{\@fpbot}{0pt plus 1fil}\makeatother

\section{Method Details}
\label{sec:supp_method}

\subsection{The Pathologist's Process}
\label{sec:supp_practice}
\Cref{tab:practice} matches each step of the pathologist's process to \method, and \cref{alg:bayespath} gives its pseudocode.
\begin{table}[H]
\centering
\footnotesize
\setlength{\tabcolsep}{3pt}
\tabno{1}\caption{\textbf{The pathologist's process and \method,} step by step, with the equations of the main paper.}
\label{tab:practice}
\vspace{-0.6em}
\begin{tabular}{@{}>{\raggedright\arraybackslash}p{0.40\linewidth}@{\hspace{5pt}}>{\raggedright\arraybackslash}p{0.58\linewidth}@{}}
\toprule
Pathologist & \method \\
\midrule
\multicolumn{2}{@{}l}{\textit{All steps}} \\
Holds a differential diagnosis & Belief $b_t$ over the candidates (\cref{eq:beliefdef}) \\
\midrule
\multicolumn{2}{@{}l}{\textit{Scan}} \\
Sweeps the slide at low power & Encoder scores every field (\cref{eq:scan}) \\
\midrule
\multicolumn{2}{@{}l}{\textit{Look}} \\
Surveys while unsure~\cite{mercan} & Survey with probability $\tilde H(b)$ (\cref{eq:survey}) \\
Tests rival diagnoses & View best separating $h_k$ from $\mathcal{D}$ (\cref{eq:look}) \\
Confirms at high power & 20$\times$ sub-field of a read field (\cref{eq:look}) \\
Describes what is seen & Observer writes observation $o_t$ \\
\midrule
\multicolumn{2}{@{}l}{\textit{Weigh}} \\
Weighs each finding & Verifier scores $o_t$ per candidate (\cref{eq:verifier}) \\
Discounts common findings & Bias-corrected ratio $\lambda_{tk}$ (\cref{eq:inversion}) \\
Knows look-alike diagnoses & Confusions $B$, learned label-free (\cref{eq:em}) \\
Revises after a finding~\cite{ledley} & Bayes update of $b_t$ (\cref{eq:belief}) \\
\midrule
\multicolumn{2}{@{}l}{\textit{Decide}} \\
Commits once it is clear & Answer once $\max_k b_t(k)\ge\tau$ (\cref{eq:decide}) \\
\bottomrule
\end{tabular}
\vspace{-0.8em}
\end{table}

\begin{algorithm}[t]
\caption{\method.}
\label{alg:bayespath}
\small
\algrenewcommand\algorithmicrequire{\textbf{Input:}}\algrenewcommand\algorithmicensure{\textbf{Output:}}
\begin{algorithmic}[1]
\Require slide $(v_0,\mathcal{F})$, question $x$, candidates $\mathcal{H}$
\algrenewcommand\algorithmicrequire{\textbf{Parameters:}}\Require $T$, $\tau$, $m$; label-free statistics $\bar q^{\mathrm{enc}}$, $\bar q$, $B$
\Ensure answer $\hat y$, belief $b_{t^\star}$, evidence trail
\Statex $\triangleright$ \textbf{Scan}
\State $b_{-1}\gets\mathrm{normalise}(\lambda^{\mathrm{scan}})$, uniform in VQA \Comment{\cref{eq:scan}}
\State $L_{-1}\gets\mathbf{1}$;\; $t\gets-1$
\While{$\max_k b_t(k)<\tau$ \textbf{and} $t<T$}
  \State $t\gets t+1$
  \Statex \hspace{\algorithmicindent}$\triangleright$ \textbf{Look}
  \If{$t=0$}
    \State $v_t\gets v_0$ \Comment{thumbnail}
  \ElsIf{$u<\tilde H(b_{t-1})$, with $u\sim\mathcal{U}(0,1)$}
    \State $v_t\sim\mathcal{U}(\text{unread fields})$ \Comment{survey, \cref{eq:survey}}
  \Else
    \State $k\sim b_{t-1}$;\; $\mathcal{D}\gets\textsc{Rivals}(k,b_{t-1},m)$ \Comment{test}
    \State $f^\star\gets\argmax_{f\in\mathcal{O}}\rho_k(f)$ \Comment{$h_k$ vs.\ $\mathcal{D}$, \cref{eq:look}}
    \If{$f^\star$ is unread}
      \State $v_t\gets f^\star$ \Comment{5$\times$}
    \Else
      \State $v_t\gets$ best unread sub-field of $f^\star$ by $\rho_k$ \Comment{20$\times$}
    \EndIf
  \EndIf
  \State $o_t\gets\textsc{Observer}(v_t,x,\mathcal{H})$
  \Statex \hspace{\algorithmicindent}$\triangleright$ \textbf{Weigh}
  \State $\lambda_t\gets\textsc{Verifier}(o_t,x,\mathcal{H})\oslash\bar q$ \Comment{\cref{eq:verifier}}
  \State $L_t\gets L^{\mathrm{conf}}_t$, or $L^{\mathrm{bound}}_t$ in VQA \Comment{\cref{eq:confusion} / (\ref{eq:update})}
  \State $b_t\gets\mathrm{normalise}\big(b_{t-1}\odot L_t\oslash L_{t-1}\big)$ \Comment{\cref{eq:belief}}
\EndWhile
\Statex $\triangleright$ \textbf{Decide}
\State $t^\star\gets t$;\; $\hat y\gets\argmax_k b_{t^\star}(k)$ \Comment{\cref{eq:decide}}
\State \Return $\hat y$, $b_{t^\star}$, $\{(v_s,o_s,b_s)\}_{s=0}^{t^\star}$
\end{algorithmic}
\end{algorithm}

\subsection{Problem Setup, Notation and Assumptions}
\label{sec:supp_math}

\begin{table*}[t]
\centering
\footnotesize
\setlength{\tabcolsep}{6pt}
\renewcommand{\arraystretch}{1.15}
\tabno{2}\caption{\textbf{Notation} by step, with the defining equation of the main paper.}
\label{tab:notation}
\vspace{-0.6em}
\begin{tabular}{@{}l@{\hspace{14pt}}l@{}}
\toprule
Notation & Definition \\
\midrule
\multicolumn{2}{@{}l}{\textit{Inputs and belief}} \\
$x$, $\mathcal{H}=\{h_1,\dots,h_K\}$ & Question and its $K$ candidate answers \\
$v_0$, $\mathcal{F}$, $f$ & Thumbnail, 5$\times$ fields of the tissue grid, one field \\
$v_t$, $o_t$ & View $t\in\{0,\dots,T\}$ and its observation \\
$T$ & View budget after the thumbnail ($T{+}1$ views in all) \\
$b_t(k)$ & Belief in $h_k$ after view $t$ (\cref{eq:beliefdef}) \\
\midrule
\multicolumn{2}{@{}l}{\textit{Scan}} \\
$e(\cdot)$, $\kappa$ & Encoder embeddings, logit scale (\cref{eq:enc}) \\
$q^{\mathrm{enc}}(k\mid f)$ & Encoder score of $h_k$ on field $f$ (\cref{eq:enc}) \\
$\bar q^{\,\mathrm{enc}}_k$ & Encoder's answer bias (\cref{eq:scan}) \\
$\lambda^{\mathrm{scan}}_k$, $b_{-1}(k)$ & Scan ratio; initial belief, uniform in VQA (\cref{eq:scan}) \\
\midrule
\multicolumn{2}{@{}l}{\textit{Look}} \\
$b$, $\tilde H(b)$ & Current belief $b_{t-1}$; its normalised entropy, the survey probability (\cref{eq:survey}) \\
$\mathcal{D}$, $m$ & $h_k$ and rivals: fewest top candidates ($\ge2$) with total belief $\ge m$ (\cref{eq:look}) \\
$\rho_k(v)$ & Encoder probability of $h_k$ within $\mathcal{D}$ on view $v$ (\cref{eq:look}) \\
$\mathcal{O}$, $f^\star$ & Fields with an unread view; best field for the test (\cref{eq:look}) \\
\midrule
\multicolumn{2}{@{}l}{\textit{Weigh}} \\
$p_{\mathrm{yes}}(o_t,h_k)$ & Verifier's probability of ``yes'' (\cref{eq:verifier}) \\
$q(k\mid o_t)$ & Normalised Verifier score (\cref{eq:verifier}) \\
$\bar q_k$ & Verifier's answer bias, uniform in VQA (\cref{eq:inversion}) \\
$\lambda_{tk}$ & Bias-corrected likelihood ratio (\cref{eq:inversion,eq:verifier}) \\
$L^{\mathrm{NB}}_{tk}$ & Naive Bayes evidence, $\prod_{s\le t}\lambda_{sk}$ \\
$L^{\mathrm{bound}}_{tk}$ & Bounded evidence (\cref{eq:update}) \\
$c_t$ & Vote of view $t$, $\argmax_k\lambda_{tk}$ \\
$B_{kj}$ & Observer's confusions, $P(c=j\mid h_k)$ (\cref{eq:em}) \\
$F_{ij}$, $W_{ik}$ & Vote share of slide $i$ for $h_j$; probability slide $i$ has answer $h_k$ (\cref{eq:em}) \\
$b^{(i)}_{-1}$ & Initial belief of slide $i$, EM's starting point (\cref{eq:em}) \\
$L^{\mathrm{conf}}_{tk}$ & Confusion-aware evidence (\cref{eq:confusion}) \\
$L_{tk}$ & Evidence used: $L^{\mathrm{conf}}$ (classification), $L^{\mathrm{bound}}$ (VQA) (\cref{eq:belief}) \\
\midrule
\multicolumn{2}{@{}l}{\textit{Decide}} \\
$\tau$, $t^\star$ & Decide threshold, stopping step (\cref{eq:decide}) \\
$\hat y$ & Predicted answer (\cref{eq:decide}) \\
\midrule
\multicolumn{2}{@{}l}{\textit{Derivations in this supplement}} \\
$p(o\mid h_k)$ & Likelihood of an observation under $h_k$ \\
$\pi$, $p_\pi(o)$ & Model's implicit prior; distribution of typical observations \\
$d$ & View carrying the slide's evidence \\
\bottomrule
\end{tabular}
\vspace{-0.6em}
\end{table*}

\Cref{tab:notation} summarises the notation. The derivations rest on four assumptions.
\begin{itemize}\itemsep1pt
\item \phantomsection\label{asm:A1}\textbf{Assumption 1 (implicit prior).} A frozen model's normalised score $q(k\mid o)$ is the Bayes posterior of $h_k$ under the model's own implicit prior $\pi$; typical observations then follow \mbox{$p_\pi(o)=\sum_j\pi_j\,p(o\mid h_j)$}.
\item \phantomsection\label{asm:A2}\textbf{Assumption 2 (multiple instances).} As in multiple-instance learning~\cite{abmil}, the $T{+}1$ views of a slide form a bag: a view $d$, drawn uniformly, carries the slide's evidence, $o_d\sim p(o\mid h_k)$, and the other views are drawn independently from $p_\pi(o)$, which does not depend on the answer.
\item \phantomsection\label{asm:A3}\textbf{Assumption 3 (two sources).} Given the answer, the scan and the Observer's views are independent; they come
from different models with different inputs.
\item \phantomsection\label{asm:A4}\textbf{Assumption 4 (reference slides).} The observations of the other slides with the same candidates are typical,
that is, distributed as $p_\pi$.
\end{itemize}

\subsection{Scan: An Initial Belief at Low Power}
\label{sec:supp_calibration}
\paragraph{Views.} \method reads three kinds of view, each sent to the Observer as an image of at most 1024 pixels per side. The \emph{thumbnail} shows the whole section, resized so that its longest side is 1024 pixels. A 5$\times$ \emph{field} is one $4096\times4096$-pixel tile of the CLAM tissue grid~\cite{clam} at full resolution, downsampled to $1024\times1024$ pixels, following~\cite{pathagent}. A 20$\times$ \emph{sub-field} is one cell of a $4\times4$ grid over a field: a $1024\times1024$-pixel tile read at full resolution, as in~\cite{hivemil}. It covers one sixteenth of the field at four times the magnification. Sub-fields with less than 50\% tissue are skipped.

\paragraph{Answer bias.}
Under \hyperref[asm:A1]{Assumption~1}, Bayes' rule gives the model's score as
\begin{equation*}
q(k\mid o)=\frac{\pi_k\,p(o\mid h_k)}{p_\pi(o)}.
\end{equation*}
Averaging over typical observations, $o\sim p_\pi$, recovers the implicit prior,
\begin{equation*}
\begin{aligned}
\mathbb{E}_{o\sim p_\pi}\big[q(k\mid o)\big]&=\int p_\pi(o)\,\frac{\pi_k\,p(o\mid h_k)}{p_\pi(o)}\,do\\
&=\pi_k\int p(o\mid h_k)\,do=\pi_k,
\end{aligned}
\end{equation*}
and dividing Bayes' rule by $\pi_k$ cancels the prior,
\begin{equation*}
\lambda_k(o)=\frac{q(k\mid o)}{\pi_k}=\frac{p(o\mid h_k)}{p_\pi(o)},
\end{equation*}
so $\lambda_k$ is a likelihood ratio with mean one, $\mathbb{E}_{o\sim p_\pi}[\lambda_k(o)]=\int p(o\mid h_k)\,do=1$. A typical observation thus adds no evidence on average, and only an observation that fits $h_k$ better than usual ($\lambda_k>1$) raises its belief. The model does not give its implicit prior, so \cref{eq:inversion} replaces $\pi_k$ with an estimate, the answer bias $\qbar_k$. By the identity above, we take $\qbar_k$ as the model's mean score over the observations of other slides with the same candidates; these are typical by \hyperref[asm:A4]{Assumption~4}, so $\qbar_k$ is a consistent estimate of $\pi_k$, obtained without labels. It is cross-fitted (\cref{sec:supp_impl}). The encoder (Scan) and the Verifier (Weigh) each have their own answer bias, computed in the same way. In slide VQA, each question has its own candidates, so the answer bias is uniform. Instead, the Observer sees the candidates in a new order for each view, which removes its preference for list positions~\cite{zheng2024pride}.

\paragraph{Scan.} The scan observes every field. By \hyperref[asm:A2]{Assumption~2}, each field is equally likely to carry the slide's evidence, so the likelihood of the scan is the average of the field ratios, $\lambda^{\mathrm{scan}}$ in \cref{eq:scan}.

\subsection{Look: Choosing the Next View}
\label{sec:supp_explorer_impl}
The Explorer chooses each view in one of two ways (\cref{eq:survey,eq:look}). A \emph{survey} reads an unread field; once every field has been read, it reads an unread 20$\times$ sub-field. A \emph{test} first updates the rivals $\mathcal{D}$ from the current belief. It then reads the best open field $f^\star$ at 5$\times$ or, if $f^\star$ has already been read, its best unread 20$\times$ sub-field.

\subsection{Look and Weigh: Describing a View}
\label{sec:supp_prompts}
The Observer uses the prompt of~\cite{pathagent}, which asks it to answer the question and list the visible features that support its answer. The Verifier's query states the question and the hypothesis that $h_k$ is the correct answer, and asks whether the observation would be expected if it were true. For negation questions, such as ``which feature is \emph{not} observed?'', the hypothesis states that the feature named by $h_k$ is absent or less evident than the others, and composite candidates are expanded into the candidates they refer to, as in~\cite{wong2026evidence}. Following~\cite{qwen3reranker}, $p_{\mathrm{yes}}=\sigma(\ell_{\mathrm{yes}}-\ell_{\mathrm{no}})$ in \cref{eq:verifier} is read from the logits $\ell$ of the tokens ``yes'' and ``no'' at the first answer position, with the sigmoid $\sigma$.

\subsection{Weigh: Deriving the Likelihood}
\label{sec:supp_update}
\paragraph{Bounded evidence (\cref{eq:update}).}
Let $o_{0:t}=(o_0,\dots,o_t)$ be the observations of the $t{+}1$ views read so far, indexed by $s\in\{0,\dots,t\}$, with $o_0$ the thumbnail; the other $T-t$ views of the budget are not read yet. By \hyperref[asm:A2]{Assumption~2}, the view $d\in\{0,\dots,T\}$ that carries the slide's evidence is drawn uniformly, its observation follows $p(o\mid h_k)$, and the observation of every other view follows $p_\pi(o)$, which does not depend on $k$. We derive the likelihood $P(o_{0:t}\mid h_k)$ in four steps.

\emph{Step 1: sum over the view that carries the evidence.} The position $d$ is uniform and independent of the answer, so by the law of total probability
\begin{equation}
P(o_{0:t}\mid h_k)=\frac{1}{T{+}1}\sum_{d=0}^{T}P(o_{0:t}\mid h_k,d).
\label{eq:s_total}
\end{equation}

\emph{Step 2: the evidence view has been read ($d\le t$).} View $d$ follows $p(o\mid h_k)$, the other read views follow $p_\pi(o)$, and the unread views integrate out. Multiplying and dividing by $p_\pi(o_d)$,
\begin{align}
P(o_{0:t}\mid h_k,d)&=p(o_d\mid h_k)\prod_{s\le t,\,s\ne d}p_\pi(o_s)\nonumber\\
&=\lambda_{dk}\prod_{s\le t}p_\pi(o_s),
\label{eq:s_given_d}
\end{align}
where $\lambda_{dk}=p(o_d\mid h_k)/p_\pi(o_d)$ is the likelihood ratio of \cref{eq:inversion} (\cref{sec:supp_calibration}).

\emph{Step 3: the evidence view has not been read ($d>t$).} Every read view is typical, and the unread views, the evidence view among them, integrate out, so
\begin{equation}
P(o_{0:t}\mid h_k,d)=\prod_{s\le t}p_\pi(o_s).
\label{eq:s_ratio_d}
\end{equation}

\emph{Step 4: average.} We split the sum in \cref{eq:s_total} into the read positions $d=0,\dots,t$, given by \cref{eq:s_given_d}, and the unread positions $d=t{+}1,\dots,T$, given by \cref{eq:s_ratio_d}. Taking out the factor $\prod_{s\le t}p_\pi(o_s)$, common to every term, gives
\begin{align*}
P(o_{0:t}\mid h_k)
&=\frac{1}{T{+}1}\Big[\sum_{d=0}^{t}\lambda_{dk}\prod_{s\le t}p_\pi(o_s)+\sum_{d=t+1}^{T}\prod_{s\le t}p_\pi(o_s)\Big]\\
&=\frac{1}{T{+}1}\Big[\sum_{d=0}^{t}\lambda_{dk}+\sum_{d=t+1}^{T}1\Big]\prod_{s\le t}p_\pi(o_s).
\end{align*}
The second sum counts the unread positions, $\sum_{d=t+1}^{T}1=T-t$. Dividing by $\prod_{s\le t}p_\pi(o_s)$ and renaming the read position $d$ as $s$,
\begin{align}
\frac{P(o_{0:t}\mid h_k)}{\prod_{s\le t}p_\pi(o_s)}
&=\frac{1}{T{+}1}\Big[\sum_{s\le t}\lambda_{sk}+(T-t)\Big]\nonumber\\
&=\frac{1}{T{+}1}\Big[\sum_{s\le t}(\lambda_{sk}-1)+(T{+}1)\Big]\nonumber\\
&=1+\frac{1}{T{+}1}\sum_{s\le t}\big(\lambda_{sk}-1\big),
\label{eq:s_bound}
\end{align}
where the second line writes $T-t=(T{+}1)-(t{+}1)$ and moves $-(t{+}1)$ into the sum as one $-1$ per ratio. This is $L^{\mathrm{bound}}_{tk}$ of \cref{eq:update}; as $\prod_{s\le t}p_\pi(o_s)$ does not depend on $k$, $L^{\mathrm{bound}}_{tk}$ is the likelihood of $h_k$ up to a constant.

\paragraph{Properties.}
Both forms of evidence depend on the views only through a sum over the views read, so they are \emph{permutation-invariant}, and an unread view adds nothing. They are also \emph{bounded}.
By the first line of \cref{eq:s_bound}, $L^{\mathrm{bound}}_{tk}$ is the average of $T{+}1$ positive numbers, $\lambda_{sk}$ for each view read and 1 for each view not yet read, so
$0<L^{\mathrm{bound}}_{tk}\le\max\big(1,\max_{s\le t}\lambda_{sk}\big)$: however many views agree, they never support $h_k$ more than its most supportive view, whereas naive Bayes, $\prod_{s\le t}\lambda_{sk}$, grows without bound.
For $L^{\mathrm{conf}}$, the logarithm of \cref{eq:confusion} is
\begin{equation}
\log L^{\mathrm{conf}}_{tk}=\frac{1}{T{+}1}\sum_{s\le t}\log B_{kc_s}.
\label{eq:s_conf}
\end{equation}
For two candidates $h_k$ and $h_{k'}$, let $\delta_s=\log(B_{kc_s}/B_{k'c_s})$ be the log-odds that vote $s$ alone would add at full weight. The $t{+}1\le T{+}1$ votes read so far change the log-odds of $h_k$ against $h_{k'}$ by
\begin{align*}
\big|\log L^{\mathrm{conf}}_{tk}-\log L^{\mathrm{conf}}_{tk'}\big|
&=\Big|\frac{1}{T{+}1}\sum_{s\le t}\delta_s\Big|\\
&\le\frac{t{+}1}{T{+}1}\max_{s\le t}|\delta_s|\le\max_{s\le t}|\delta_s|,
\end{align*}
so, as with $L^{\mathrm{bound}}$, the votes never shift the log-odds more than their most informative vote would alone. In both forms, the weight $1/(T{+}1)$ is set by the view budget, not tuned.

\begin{figure*}[t]
\centering
\includegraphics[width=0.9\textwidth]{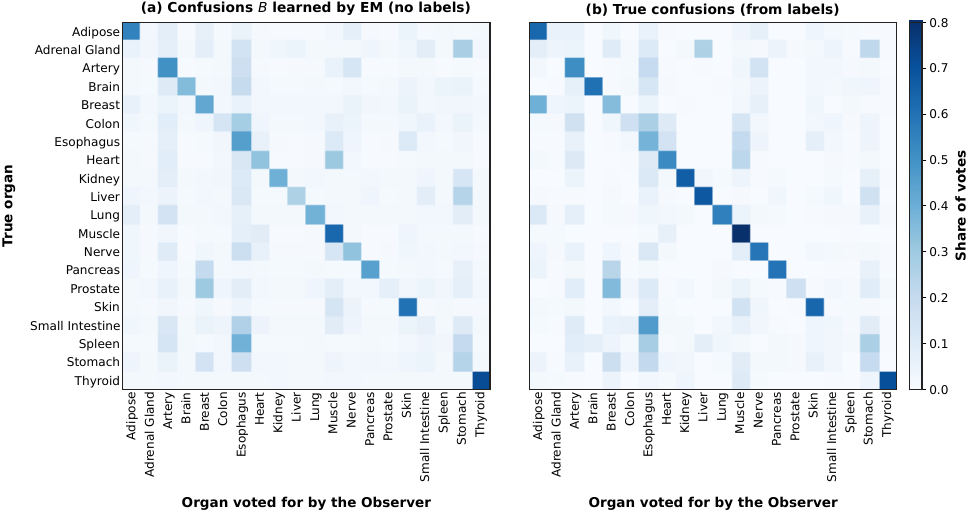}
\vspace{-0.4em}
\caption{\textbf{The Observer's confusions on GTEx.} (a)~Learned by EM without labels (\cref{eq:em}). (b)~True confusions of the same votes (evaluation only). Rows: true organ; columns: organ voted for. EM recovers the main misreadings, \eg heart read as muscle.}
\label{fig:confusion}
\vspace{-0.6em}
\end{figure*}
\paragraph{The belief update and Bayes' rule.}
The main paper obtains \cref{eq:belief} from $b_t(k)\propto b_{-1}(k)\,L_{tk}$, where the scan prior and the evidence of the views multiply by \hyperref[asm:A3]{Assumption~3}. With $L^{\mathrm{bound}}$, the likelihood up to a constant, this is Bayes' rule, and so is each step of \cref{eq:belief}: by \cref{eq:s_bound} and the chain rule,
\begin{align}
\frac{L^{\mathrm{bound}}_{tk}}{L^{\mathrm{bound}}_{t-1,k}}
&=\frac{P(o_{0:t}\mid h_k)}{P(o_{0:t-1}\mid h_k)}\cdot\frac{\prod_{s\le t-1}p_\pi(o_s)}{\prod_{s\le t}p_\pi(o_s)}\nonumber\\
&=\frac{P(o_t\mid o_{0:t-1},h_k)}{p_\pi(o_t)}.
\label{eq:s_exact}
\end{align}
As $p_\pi(o_t)$ does not depend on $k$, the factor of each view is, up to a constant, the likelihood of the new observation given the earlier ones: under \hyperref[asm:A1]{Assumptions~1}--\hyperref[asm:A3]{3}, with $\bar q_k=\pi_k$, \cref{eq:belief} is exact Bayes' rule. With $L^{\mathrm{conf}}$, the sums in \cref{eq:s_conf} for $t$ and $t{-}1$ differ only by the new vote, so the factor is $B_{kc_t}^{1/(T+1)}$, the likelihood of the vote raised to the power $1/(T{+}1)$: the tempered Bayes update of~\cite{bissiri2016general}, under which the correlated votes of one slide together count as only one observation, as in \cref{eq:confusion}.

\subsection{Weigh: Estimating the Confusions}
\label{sec:supp_habits}
\paragraph{Derivation of the EM updates (\cref{eq:em}).}
\Cref{eq:em} adapts the Dawid--Skene model~\cite{dawidskene}, which estimates the confusion matrices of annotators from their labels alone, without ground truth, and allows an annotator to label an item several times. Here the annotator is the Observer, and it labels each slide once per view: slide $i$ has $n_i$ votes, of which $n_{ij}$ are for $h_j$, so $F_{ij}=n_{ij}/n_i$; its answer $z_i$ is never observed. We derive \cref{eq:em} in four steps.

\emph{Step 1: the model.} We make three changes to Dawid--Skene. (i) The prior of $z_i$ is the scan belief of slide $i$, $\beta_{ik}=b^{(i)}_{-1}(k)$, rather than a class prior shared by all slides. (ii) Given $z_i=k$, each vote is $j$ with probability $B_{kj}$, and, as in \cref{eq:confusion}, the votes of a slide count as one observation: each enters with the power $1/n_i$, the counterpart of $1/(T{+}1)$ once all views of the slide are read. The $n_{ij}$ votes for $j$ then give $B_{kj}^{n_{ij}/n_i}=B_{kj}^{F_{ij}}$, so the tempered likelihood of slide $i$ under $z_i=k$ is
\begin{equation}
\tilde p_{ik}(B)=\prod_j B_{kj}^{F_{ij}}.
\end{equation}
(iii) Each row of $B$ has the prior $\mathrm{Dir}(1+\frac{1}{K},\dots,1+\frac{1}{K})$, whose log density is $\frac{1}{K}\sum_j\log B_{kj}$ up to a constant.

\emph{Step 2: the objective.} Summing over the unknown answers and adding the prior, EM maximises, under $\sum_j B_{kj}=1$ for every $k$, the tempered log posterior
\begin{equation}
\mathcal{J}(B)=\sum_i\log\sum_k\beta_{ik}\,\tilde p_{ik}(B)+\frac{1}{K}\sum_{k,j}\log B_{kj}.
\label{eq:s_emobj}
\end{equation}

\emph{Step 3: the E-step.} EM maximises a lower bound on \cref{eq:s_emobj}. For any distribution $W_{i\cdot}$ over the answers of slide $i$, we multiply and divide each term by $W_{ik}$ and apply Jensen's inequality, as the logarithm is concave:
\begin{equation}
\begin{aligned}
&\log\textstyle\sum_k\beta_{ik}\,\tilde p_{ik}(B)
=\log\textstyle\sum_k W_{ik}\,\frac{\beta_{ik}\tilde p_{ik}(B)}{W_{ik}}\\
&\quad\ge\textstyle\sum_k W_{ik}\log\frac{\beta_{ik}\tilde p_{ik}(B)}{W_{ik}}.
\end{aligned}
\label{eq:s_jensen}
\end{equation}
The E-step makes the bound tight: equality holds when the ratio $\beta_{ik}\tilde p_{ik}(B)/W_{ik}$ is the same for every $k$, so, normalising over $k$,
\begin{equation}
W_{ik}=\frac{\beta_{ik}\,\tilde p_{ik}(B)}{\sum_{k'}\beta_{ik'}\,\tilde p_{ik'}(B)},
\label{eq:s_estep}
\end{equation}
the tempered posterior of $z_i$ under the current $B$. As $\beta_{ik}=b^{(i)}_{-1}(k)$ and $\tilde p_{ik}(B)=\exp(\sum_j F_{ij}\log B_{kj})$, this is $W_{ik}\propto b^{(i)}_{-1}(k)\exp(\sum_j F_{ij}\log B_{kj})$, the E-step of \cref{eq:em}.

\emph{Step 4: the M-step.} With $W$ fixed by \cref{eq:s_estep}, we sum \cref{eq:s_jensen} over the slides and add the prior term of \cref{eq:s_emobj}. As $\log\tilde p_{ik}(B)=\sum_j F_{ij}\log B_{kj}$, the terms that depend on $B$ are
\begin{equation*}
Q(B)=\sum_{k,j}\gamma_{kj}\log B_{kj},\qquad \gamma_{kj}=\frac{1}{K}+\sum_i W_{ik}F_{ij}.
\end{equation*}
With one Lagrange multiplier $\nu_k$ for each constraint $\sum_j B_{kj}=1$, setting the derivative to zero gives $\gamma_{kj}/B_{kj}=\nu_k$, so $B_{kj}\propto \gamma_{kj}=\frac{1}{K}+\sum_i W_{ik}F_{ij}$, the M-step of \cref{eq:em}; as $Q$ is concave, this is the maximum. The prior adds one pseudo-slide with uniform votes, $\frac{1}{K}$ each, to every row, so a rarely seen answer keeps a nearly uniform row and every $B_{kj}>0$.

\emph{Convergence.} The E-step makes the bound equal to $\mathcal{J}(B)$ and the M-step raises it, so $\mathcal{J}$ never decreases~\cite{dempster1977em}. Without labels, the votes alone would determine the rows of $B$ only up to a permutation of the answers; the scan belief, used as the prior and as the starting point $W_{ik}=b^{(i)}_{-1}(k)$, ties each row to its answer.

\paragraph{The estimated confusions.} \Cref{fig:confusion} compares $B$ for GTEx with the true confusions of the same votes: EM recovers the Observer's main misreadings, such as heart read as muscle and spleen, colon or small intestine as esophagus.

\subsection{Decide: When to Answer}
\label{sec:supp_signout}
Decide answers once the belief reaches $\tau$, and Look chooses each view from the current belief, so the views are chosen adaptively. Neither requires a correction to the belief. Let $D_{t-1}$ be the data observed before view $t$: the scan and the views and observations $v_{0:t-1}$, $o_{0:t-1}$. The Explorer draws $v_t$ from a policy $\mu(v_t\mid D_{t-1})$ that depends only on $D_{t-1}$; its own randomness (the survey and the draw $k\sim b_{t-1}$) is independent of the answer. The probability of the views and observations under $h_k$ therefore factorises as
\begin{equation*}
\prod_{t}\mu(v_t\mid D_{t-1})\;P(o_t\mid v_t,D_{t-1},h_k).
\end{equation*}
The policy factors do not depend on $k$ and cancel when the belief is normalised, so \cref{eq:belief} needs no correction for the adaptive choice of views. For the same reason, stopping once $\max_k b_t(k)\ge\tau$ needs no correction, as the stopping rule also depends only on $D_t$.

\begin{table*}[t]
\centering
\footnotesize
\setlength{\tabcolsep}{2pt}
\tabno{4}\caption{\textbf{Training-free WSI agents} of \cref{tab:main}: frozen models and stopping rule. Belief: \method's belief over the candidates.}
\label{tab:supp_methods}
\vspace{-0.6em}
\begin{tabular}{@{}lllll@{}}
\toprule
Agent & Choose views & Describe views & Combine observations & Stop \\
\midrule
GIANT~\cite{giant} & Patho-R1-7B & Patho-R1-7B & Patho-R1-7B & MLLM decides \\
PathAgent~\cite{pathagent} & CONCH & Patho-R1-7B & Qwen3-4B & LLM judges sufficiency \\
PathNavigate~\cite{pathnavigate} & CONCH & Patho-R1-7B & Qwen3-4B & fixed budget of 15 views \\
\rowcolor{cbblue!7}\method & CONCH + belief & Patho-R1-7B & Qwen3-Reranker-4B + belief & belief $\ge\tau$ \\
\bottomrule
\end{tabular}
\vspace{-0.6em}
\end{table*}

\section{Experimental Setup}
\label{sec:supp_exp}

\subsection{Datasets and Metrics}
\label{sec:supp_tasks}
\Cref{tab:supp_data} lists the five tasks of MultiPathQA~\cite{giant}. All slides come from public, de-identified releases and are used under their terms of use. The expected calibration error (ECE, \cref{tab:trust}) uses ten equal-width bins of the confidence of the answer, $\max_k b_{t^\star}(k)$. AURC takes the error rate of the kept answers as risk; LLM aggregation states no confidence, so its answers are ranked by the share of agreeing observations.

\begin{table}[H]
\centering
\footnotesize
\setlength{\tabcolsep}{3pt}
\tabno{3}\caption{\textbf{Tasks}: questions, slides, candidates $K$ and metric.}
\label{tab:supp_data}
\vspace{-0.6em}
\resizebox{\linewidth}{!}{%
\begin{tabular}{@{}llcccc@{}}
\toprule
Task & Question & $K$ & Questions & Slides & Metric \\
\midrule
TCGA~\cite{tcga} & cancer type & 30 & 221 & 221 & BACC \\
GTEx~\cite{gtex} & organ of normal tissue & 20 & 190 & 190 & BACC \\
PANDA~\cite{panda} & ISUP grade & 6 & 196 & 196 & BACC \\
ExpertVQA~\cite{giant} & pathologist-written & 4 & 128 & 76 & ACC \\
SlideBench~\cite{slidechat} & SlideBench-VQA (TCGA) & 4 & 197 & 183 & ACC \\
\bottomrule
\end{tabular}}
\vspace{-0.8em}
\end{table}

\subsection{Baselines}
\label{sec:supp_models}
\paragraph{MLLMs.} The general and medical MLLMs see the slide thumbnail (longest side 1024 pixels), the question and the lettered candidates, and answer with one letter. One parser maps every reply to a candidate; a reply that names none counts as wrong. Replies are sampled at temperature 0.7, as for the Observer, and each model is run with three seeds.
\paragraph{Supervised WSI models.} WSI-VQA~\cite{wsivqa}, SlideChat~\cite{slidechat} and WSI-LLaVA~\cite{wsillava} run with their released checkpoints and official inference code, on slide features of the encoder each was trained with (ResNet-50, CONCH and Prov-GigaPath). SlideChat and WSI-LLaVA receive the same prompt as the MLLMs. WSI-VQA receives the question alone, as in its official evaluation, and its answer is mapped to the closest candidate by its official close-ended matching.
\paragraph{Training-free WSI agents.} \Cref{tab:supp_methods} lists, for each agent, the models that choose the views, describe them and combine the observations, and its stopping rule.

\subsection{Parameters}
\label{sec:supp_params}
\Cref{tab:supp_params} lists the parameters of \method.
\begin{table}[t]
\centering
\footnotesize
\setlength{\tabcolsep}{3pt}
\tabno{5}\caption{\textbf{Parameters of \method,} shared by all five tasks.}
\label{tab:supp_params}
\vspace{-0.6em}
\begin{tabular}{@{}lc@{}}
\toprule
Parameter & Value \\
\midrule
View budget $T$ (after the thumbnail) & 25 \\
Decide threshold $\tau$ & 0.95 \\
Minimum belief mass of the rivals $\mathcal{D}$, $m$ & 0.95 \\
Field (level 0) / image sent to the Observer & 4096 / 1024 pixels \\
Sub-fields per field & $4\times4$ \\
Minimum tissue in a sub-field & 50\% \\
Observer sampling: temperature / top-$p$ & 0.7 / 0.95 \\
Runs (seeds) & 3 \\
EM iterations & 50 \\
Folds of the cross-fitting & 2 \\
\bottomrule
\end{tabular}
\vspace{-0.8em}
\end{table}

\paragraph{Confidence threshold.}\label{sec:supp_anytime} \method stops reading once its belief in one answer reaches $\tau$; we use $\tau=0.95$. To test whether this choice matters, we take the belief recorded after every view of each run and, for each $\tau$, read off the answer at the first view where $\max_k b_t(k)\ge\tau$. Every $\tau$ from 0.7 to 0.99 gives the same answer on every question of all five tasks. Once the belief is confident, further views do not change the answer, so \cref{tab:main} does not depend on $\tau$ in this range.

\subsection{Variants of the Ablations}
\label{sec:supp_variants}
\Cref{tab:supp_variants} writes each ablation variant of \cref{tab:memory} as an equation.
\begin{table}[t]
\centering
\footnotesize
\setlength{\tabcolsep}{3pt}
\renewcommand{\arraystretch}{1.15}
\tabno{6}\caption{\textbf{Ablation of \method as equations,} rows as in \cref{tab:memory}. Bayes' rule computes $b_T(k)\propto b_{-1}(k)\cdot\text{Evidence}$, with $L^{\mathrm{bound}}$ and $L^{\mathrm{conf}}$ from \cref{eq:update,eq:confusion}; LLM aggregation answers from the observations directly. BC: bias correction. BE: bounded evidence. CE: confusion-aware evidence. Views: belief-guided (\cref{eq:look}) or by similarity $\cos(e(v),e(x))$ to the question.}
\label{tab:supp_variants}
\vspace{-0.6em}
\begin{tabular}{@{}lccclc@{}}
\toprule
Combine & BC & BE & CE & Evidence & Views \\
\midrule
Qwen3-4B & & & & \multirow{2}{*}{$\hat y=\mathrm{LLM}(x,\mathcal{H},b_{-1},o_{0:T})$} & belief \\
Qwen3.8-27B & & & & & belief \\
\midrule
Bayes' rule & \ding{51} & & & $\prod_{t}\lambda_{tk}$ (naive Bayes) & belief \\
Bayes' rule & \ding{51} & \ding{51} & & $L^{\mathrm{bound}}_{Tk}$ & belief \\
Bayes' rule & \ding{51} & \ding{51} & \ding{51} & $L^{\mathrm{conf}}_{Tk}$; $L^{\mathrm{bound}}_{Tk}$ in slide VQA & similarity \\
\rowcolor{cbblue!7}\method & \ding{51} & \ding{51} & \ding{51} & $L^{\mathrm{conf}}_{Tk}$; $L^{\mathrm{bound}}_{Tk}$ in slide VQA & belief \\
\bottomrule
\end{tabular}
\vspace{-0.6em}
\end{table}

\subsection{Implementation}
\label{sec:supp_impl}
\paragraph{Cross-fitting.} Each task's slides are split into two folds by a hash of the slide identifier. The answer biases of the Verifier and the encoder, and $B$, are computed for each fold from the unlabelled slides of the other fold only, so no slide contributes to the statistics used for its own prediction.
\paragraph{Compute.} All models are served with vLLM on NVIDIA RTX A5000/A6000 GPUs.

\begin{figure*}[t]
\centering
\includegraphics[width=0.86\textwidth]{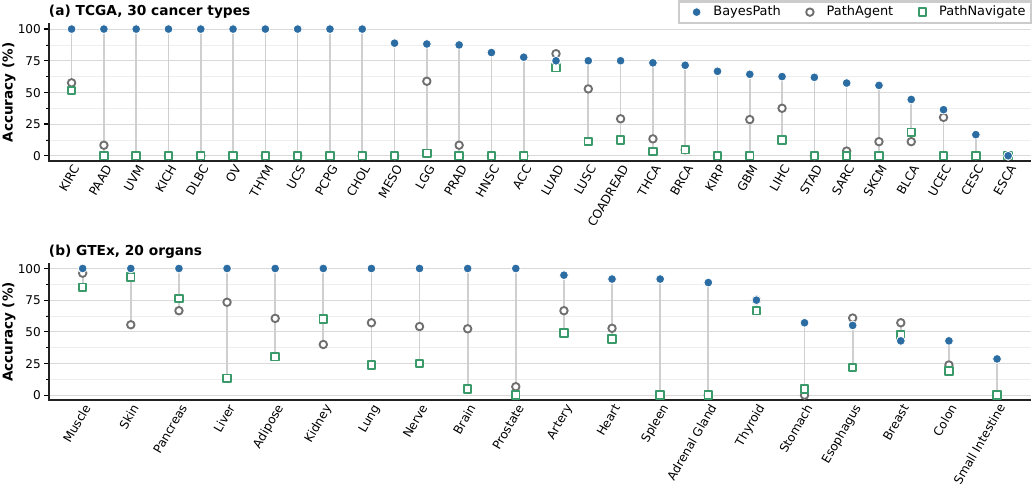}
\vspace{-0.5em}
\caption{\textbf{Accuracy on each class} (three seeds pooled), sorted by the accuracy of \method. \method matches or exceeds both agents on 29 of 30 TCGA and 18 of 20 GTEx classes. TCGA classes are labelled by project code.}
\label{fig:percls}
\vspace{-0.6em}
\end{figure*}
\section{Comparison with Existing Methods}
\label{sec:supp_cmp}

\subsection{Statistical Tests}
\label{sec:supp_stats}
\Cref{tab:supp_stats} compares \method on the same slides with PathAgent, PathNavigate, LLM aggregation and naive Bayes. Every 95\% confidence interval lies above zero, so every gain in the table is significant at the 5\% level. On TCGA and GTEx, McNemar's test gives $p<10^{-10}$ against both agents.

\input{sec/X_tab_stats}

\subsection{Where the Answers Go}
\label{sec:supp_spread}

With 20 and 30 candidates, the agents that read observations as text put most of their answers on a few default diagnoses (\cref{tab:spread}): PathAgent and PathNavigate answer lung adenocarcinoma for 49.6\% and 54.1\% of the TCGA slides, the label of 5.4\%. \method uses every candidate and answers 97\% of the TCGA classes and all GTEx classes correctly at least once. \Cref{fig:percls} shows the accuracy on each class.

\begin{table}[t]
\centering
\footnotesize
\setlength{\tabcolsep}{3pt}
\tabno{8}\caption{\textbf{Where the answers go} (three seeds pooled). Used: candidates ever predicted. Top-3: answers on the three most frequent candidates. Hit: classes answered correctly at least once.}
\label{tab:spread}
\vspace{-0.7em}
\begin{tabular}{@{}lccccccc@{}}
\toprule
 & \multicolumn{3}{c}{TCGA ($K{=}30$)} & & \multicolumn{3}{c}{GTEx ($K{=}20$)} \\
\cmidrule{2-4}\cmidrule{6-8}
Method & Used & Top-3 & Hit & & Used & Top-3 & Hit \\
\midrule
PathAgent & 22 & 69\% & 50\% & & 19 & 45\% & 80\% \\
PathNavigate & 17 & 89\% & 30\% & & 18 & 45\% & 80\% \\
\rowcolor{cbblue!7}\method & \textbf{30} & \textbf{21\%} & \textbf{97\%} & & \textbf{20} & \textbf{29\%} & \textbf{100\%} \\
\bottomrule
\end{tabular}
\vspace{-0.8em}
\end{table}

\section{The Role of the Belief}
\label{sec:supp_bayes}

\subsection{The Belief Combines the Evidence}
\label{sec:supp_same}
\Cref{tab:supp_same_tcga,tab:supp_same_gtex} extend \cref{tab:memory} with calibration metrics. The raw product is Bayes' rule on the uncorrected scores, $b_T(k)\propto b_{-1}(k)\prod_t q(k\mid o_t)$. \method is the most accurate and best calibrated rule on both tasks, and it is more confident when correct than when wrong.

\input{sec/X_tab_same}

\subsection{Keeping a Belief Inside the LLM}
\label{sec:supp_prompted}
Can an LLM keep the belief itself? On the evidence of \cref{tab:memory}, the belief prompt asks the LLM to keep and update a probability over the candidates after each view; the reliability prompt also warns that the reports repeat mistakes (\cref{fig:prompts}). Neither prompt closes the gap to \method on average, even with the ${\sim}7\times$ larger Qwen3.8-27B (\cref{tab:supp_prompted}).

\begin{figure}[t]
\centering
\newcommand{\ph}[1]{\textrm{\textit{\{#1\}}}}%
\newcommand{\hd}[2]{\textbf{#1}\hfill{\footnotesize #2}}%
\footnotesize
\begin{tabular}{@{}>{\raggedright\arraybackslash\scriptsize\ttfamily}p{\dimexpr\linewidth\relax}@{}}
\toprule
\normalfont\footnotesize\hd{System prompt}{both prompts} \\
\midrule
You are an expert pathologist. You are given the reports written while examining one H\&E whole-slide image, view by view, and a multiple-choice question about the slide. Keep an explicit probability over the options. Start from the low-power scan if one is given, otherwise from equal probabilities. Go through the views in order and, after each view, update the probabilities of the options it bears on, in one short line. Finish with exactly one line: FINAL: <option letter> <probability of that option, between 0 and 1> \\
\midrule
\normalfont\footnotesize\hd{Added to the system prompt}{reliability prompt only} \\
\midrule
The reports come from one imperfect model looking at one specimen. It tends to repeat the same mistake in view after view and to overuse some diagnoses, so the views are not independent: repeated mentions of a diagnosis must not be counted as separate evidence. \\
\midrule
\normalfont\footnotesize\hd{User message}{both prompts} \\
\midrule\relax
\normalfont\footnotesize\setlength{\tabcolsep}{0pt}\renewcommand{\arraystretch}{1.05}%
\begin{tabular}{@{}>{\raggedright\arraybackslash\scriptsize\ttfamily}p{\linewidth}@{}}
[Low-power scan]$^{\dagger}$ A pathology encoder (CONCH, zero-shot) scored all tissue fields of the slide. Probability of the most likely options: \\
- \ph{option}: \ph{probability} \\[3pt]\relax
[View 1] \ph{observation $o_0$} \\\relax
[View 2] \ph{observation $o_1$} \\
\ldots \\\relax
[View $T{+}1$] \ph{observation $o_T$} \\[3pt]
Question: \ph{question} \\
Options: \\
A. \ph{option 1} \\
B. \ph{option 2} \\
\ldots \\
Answer: \\
\end{tabular} \\
\bottomrule
\end{tabular}
\vspace{-0.4em}
\caption{\textbf{Belief and reliability prompts} for LLM aggregation (\cref{tab:supp_prompted}). $o_0$: observation of the thumbnail; $T$: view budget. $^{\dagger}$Slide classification only; it lists the five most likely options.}
\label{fig:prompts}
\end{figure}

\begin{table}[H]
\vspace{-0.9em}
\centering
\footnotesize
\setlength{\tabcolsep}{3pt}
\tabno{11}\caption{\textbf{Keeping a belief inside the LLM} (evidence of \cref{tab:memory}). Mean of three seeds (\%). EVQA: ExpertVQA; SB: SlideBench.}
\label{tab:supp_prompted}
\vspace{-0.6em}
\resizebox{\linewidth}{!}{%
\begin{tabular}{@{}lcccccc@{}}
\toprule
Rule & TCGA & GTEx & PANDA & EVQA & SB & Avg. \\
\midrule
Qwen3-4B aggregation & 24.8 & 55.1 & 21.5 & 43.8 & 55.7 & 40.2 \\
\quad belief prompt & 21.9 & 51.2 & 20.2 & 45.6 & 53.1 & 38.4 \\
\quad reliability prompt & 22.6 & 53.3 & 24.0 & 44.8 & 54.1 & 39.8 \\
\midrule
Qwen3.8-27B aggregation & 36.8 & 68.0 & 22.1 & \textbf{48.7} & 55.8 & 46.3 \\
\quad belief prompt & 44.6 & 66.8 & 20.3 & 48.4 & 54.8 & 47.0 \\
\quad reliability prompt & 47.7 & 68.7 & 21.7 & 47.7 & 55.3 & 48.2 \\
\midrule
\rowcolor{cbblue!7}\method & \textbf{75.3} & \textbf{83.4} & \textbf{28.2} & 47.7 & \textbf{58.4} & \textbf{58.6} \\
\bottomrule
\end{tabular}}
\vspace{-0.8em}
\end{table}

\subsection{Fewer Candidate Answers}
\label{sec:supp_fewer}
Does the lead of \method come only from the many candidates of TCGA and GTEx? To test this, we keep the slides and observations of these tasks but give each question only $K$ candidates: the true answer and $K{-}1$ random distractors, fixed per question. \method is the most accurate at every $K$ (\cref{tab:fewer}).
\begin{table}[t]
\centering
\footnotesize
\setlength{\tabcolsep}{3.2pt}
\tabno{12}\caption{\textbf{Fewer candidates} on the evidence of \cref{tab:memory}: the true answer and $K{-}1$ random distractors; the last column of each task has all candidates. Mean of three seeds (BACC, \%).}
\label{tab:fewer}
\begin{tabular}{@{}lccccccc@{}}
\toprule
 & \multicolumn{4}{c}{TCGA} & \multicolumn{3}{c}{GTEx} \\
\cmidrule(lr){2-5}\cmidrule(l){6-8}
Rule & $K{=}4$ & 10 & 20 & 30 & $K{=}4$ & 10 & 20 \\
\midrule
LLM aggregation & 65.3 & 47.6 & 33.2 & 24.8 & 77.1 & 62.9 & 55.1 \\
Naive Bayes & 68.5 & 50.0 & 41.6 & 36.0 & 88.0 & 73.6 & 65.9 \\
\rowcolor{cbblue!7}\method & \textbf{92.6} & \textbf{83.5} & \textbf{77.8} & \textbf{75.3} & \textbf{94.4} & \textbf{85.8} & \textbf{83.4} \\
\midrule
Margin over LLM & +27.3 & +35.9 & +44.7 & +50.5 & +17.3 & +22.9 & +28.3 \\
\bottomrule
\end{tabular}
\end{table}

\subsection{Reversed View Order}
\label{sec:supp_order}
The answer should depend on what the views show, not on the order in which they happen to be read. Bayes' rule is order-invariant: reading the same observations in reverse changes none of \method's answers, unlike those of LLM aggregation (\cref{tab:order}).
\begin{table}[H]
\centering
\footnotesize
\setlength{\tabcolsep}{2.1pt}
\tabno{13}\caption{\textbf{Reversed view order.} Answers that change when the observations are read in reverse; $\downarrow$: lower is better. Mean of three seeds (\%). EVQA: ExpertVQA; SB: SlideBench.}
\label{tab:order}
\vspace{-0.6em}
\begin{tabular}{@{}lccccc@{}}
\toprule
Rule & TCGA\,$\downarrow$ & GTEx\,$\downarrow$ & PANDA\,$\downarrow$ & EVQA\,$\downarrow$ & SB\,$\downarrow$ \\
\midrule
LLM aggr. (Qwen3-4B) & 31.7 & 31.1 & 32.7 & 26.1 & 20.8 \\
LLM aggr. (Qwen3.8-27B) & 7.2 & 9.7 & 26.1 & 13.0 & 13.1 \\
Naive Bayes & \textbf{0.0} & \textbf{0.0} & \textbf{0.0} & \textbf{0.0} & \textbf{0.0} \\
\rowcolor{cbblue!7}\method & \textbf{0.0} & \textbf{0.0} & \textbf{0.0} & \textbf{0.0} & \textbf{0.0} \\
\bottomrule
\end{tabular}
\vspace{-0.8em}
\end{table}

\subsection{The Belief Chooses Where to Look}
\label{sec:supp_zoom}
At 20$\times$, the views chosen by the belief point to the true answer more often than question-similarity views on every task (\cref{tab:zoom}), well above chance.

\begin{table}[H]
\centering
\footnotesize
\setlength{\tabcolsep}{2.6pt}
\tabno{14}\caption{\textbf{Belief-guided views} at 20$\times$ (three seeds pooled, \%): share of views whose observation points to the true answer. Similarity views rank the 20$\times$ views by similarity to the question (\cref{tab:memory}). Labels are used only for this analysis.}
\label{tab:zoom}
\resizebox{\linewidth}{!}{%
\begin{tabular}{@{}lcccc@{}}
\toprule
20$\times$ views & TCGA & GTEx & ExpertVQA & SlideBench \\
\midrule
\textcolor{cbgray}{chance ($1/K$)} & \textcolor{cbgray}{3.3} & \textcolor{cbgray}{5.0} & \textcolor{cbgray}{25.0} & \textcolor{cbgray}{25.0} \\
similarity views & 23.8 & 39.6 & 32.2 & 50.5 \\
\rowcolor{cbblue!7}\method (belief) & \textbf{40.4} & \textbf{47.5} & \textbf{35.1} & \textbf{59.9} \\
\bottomrule
\end{tabular}}
\end{table}

\subsection{The Belief Knows Which Answers to Trust}
\label{sec:supp_decide}
On every task, the most confident answers of \method are more accurate than those of naive Bayes (\cref{tab:decide}); on TCGA, its 20\% most confident answers are all correct.

\begin{table}[b]
\centering
\footnotesize
\setlength{\tabcolsep}{3pt}
\tabno{15}\caption{\textbf{Accuracy of the most confident answers,} ranked by $\max_k b(k)$. Mean of three seeds (\%). EVQA: ExpertVQA; SB: SlideBench; NB: naive Bayes; BP: \method.}
\label{tab:decide}
\resizebox{\linewidth}{!}{%
\begin{tabular}{@{}l>{\columncolor{white}}c>{\columncolor{cbblue!7}}c>{\columncolor{white}}c>{\columncolor{cbblue!7}}c>{\columncolor{white}}c>{\columncolor{cbblue!7}}c>{\columncolor{white}}c>{\columncolor{cbblue!7}}c>{\columncolor{white}}c>{\columncolor{cbblue!7}}c@{}}
\toprule
 & \multicolumn{2}{c}{TCGA} & \multicolumn{2}{c}{GTEx} & \multicolumn{2}{c}{PANDA} & \multicolumn{2}{c}{EVQA} & \multicolumn{2}{c}{SB} \\
\cmidrule(lr){2-3}\cmidrule(lr){4-5}\cmidrule(lr){6-7}\cmidrule(lr){8-9}\cmidrule(l){10-11}
Most confident & NB & BP & NB & BP & NB & BP & NB & BP & NB & BP \\
\midrule
10\% & 93.3 & \textbf{100.0} & 87.1 & \textbf{89.5} & 23.3 & \textbf{30.0} & 59.3 & \textbf{82.1} & 68.1 & \textbf{91.7} \\
20\% & 89.4 & \textbf{100.0} & 87.5 & \textbf{94.7} & 21.4 & \textbf{30.8} & 59.0 & \textbf{70.5} & 68.1 & \textbf{84.6} \\
30\% & 80.8 & \textbf{96.5} & 87.5 & \textbf{94.7} & 23.2 & \textbf{28.8} & 59.2 & \textbf{67.5} & 68.1 & \textbf{81.4} \\
\bottomrule
\end{tabular}}
\end{table}

\section{Reliability-Aware Likelihood}
\label{sec:supp_reliab}

\subsection{The Observer Repeats Its Mistakes}
\label{sec:supp_repeat}
Before bias correction, the votes concentrate on a few answers: on TCGA, 44.4\% vote for lung adenocarcinoma, the label of 5.4\% of slides, and on GTEx, 20.5\% vote for muscle, the label of 4.7\%. The same wrong answer also repeats within a slide (\cref{tab:repeat}). From these same observations, \method reaches 75.3\% on TCGA and 83.4\% on GTEx.

\begin{table}[H]
\centering
\footnotesize
\setlength{\tabcolsep}{4pt}
\tabno{16}\caption{\textbf{The Observer repeats its mistakes.} Each view votes for the answer the Verifier scores highest. Slides: share of slides on which at least three views vote for the same wrong answer (\%). Views: average number of views per slide that vote for the most frequent wrong answer. Mean of three seeds.}
\label{tab:repeat}
\begin{tabular}{@{}lccccc@{}}
\toprule
 & TCGA & GTEx & PANDA & ExpertVQA & SlideBench \\
\midrule
Slides (\%) & 90.0 & 80.5 & 95.1 & 93.2 & 87.3 \\
Views & 11.7 & 7.6 & 9.6 & 10.4 & 9.1 \\
\bottomrule
\end{tabular}
\end{table}

\subsection{Parts of the Likelihood}
\label{sec:supp_parts}
\Cref{tab:update} adds the parts of the likelihood in turn. Each part improves every task.

\begin{table}[H]
\centering
\footnotesize
\setlength{\tabcolsep}{3.2pt}
\tabno{17}\caption{\textbf{Parts of the likelihood} on slide classification (BACC, \%). BC: bias correction. BE: bounded evidence (a). CM: confusion matrix, \ie confusion-aware evidence (b). Mean of three seeds.}
\label{tab:update}
\begin{tabular}{@{}lcccccc@{}}
\toprule
Likelihood & BC & BE & CM & TCGA & GTEx & PANDA \\
\midrule
Raw product & & & & 20.9 & 54.6 & 21.8 \\
Naive Bayes & \ding{51} & & & 36.0 & 65.9 & 21.9 \\
Bounded evidence (a) & \ding{51} & \ding{51} & & 73.0 & 79.8 & 26.2 \\
\rowcolor{cbblue!7}\method (b) & \ding{51} & \ding{51} & \ding{51} & \textbf{75.3} & \textbf{83.4} & \textbf{28.2} \\
\bottomrule
\end{tabular}
\end{table}

\subsection{Label-Free Statistics}
\label{sec:supp_br}
The answer biases and $B$ need only unlabelled slides with the same candidates. We test two sources (\cref{tab:supp_baserate}): the other half of the test slides (default, 75.3 BACC on TCGA) and 253 TCGA slides of the slide VQA tasks, which do not overlap with the classification set (74.2).

\input{sec/X_tab_br}

\begin{figure*}[t]
\centering
\includegraphics[width=0.94\textwidth]{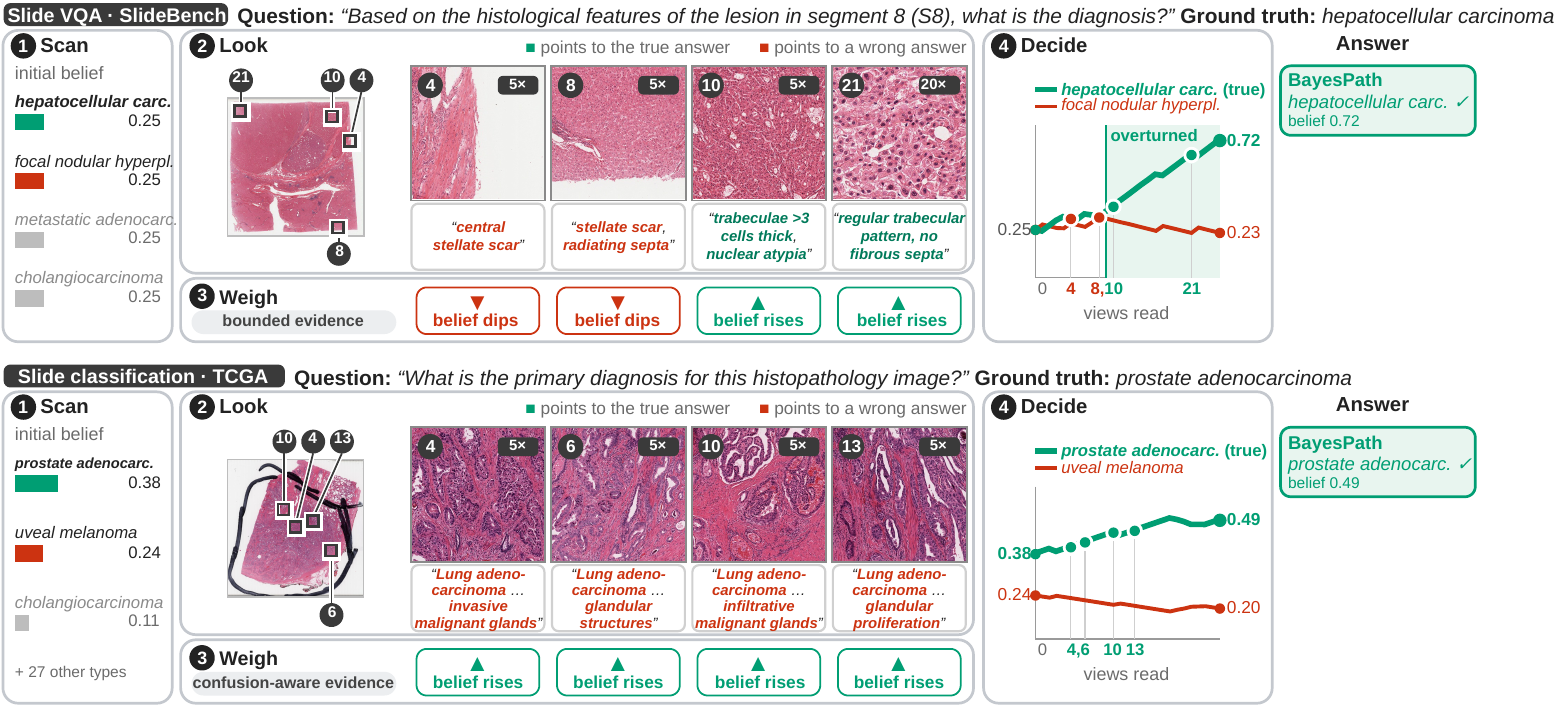}
\caption{\textbf{More evidence trails of \method.} \textbf{Top, bounded evidence}: views read as a stellate scar, favouring \textcolor[HTML]{CC3311}{focal nodular hyperplasia}, barely lower \textcolor[HTML]{007A5A}{hepatocellular carcinoma}, while views of malignant features (trabeculae more than three cells thick, nuclear atypia) and a 20$\times$ view without fibrous septa raise it (0.25$\to$0.72). \textbf{Bottom, confusion-aware evidence}: most views are read as \textcolor[HTML]{CC3311}{lung adenocarcinoma}. The label-free confusions show that views of \textcolor[HTML]{007A5A}{prostate adenocarcinoma} slides vote lung adenocarcinoma more often (60\%) than views of lung adenocarcinoma (54\%) or uveal melanoma slides (25\%), so these votes raise prostate adenocarcinoma (0.38$\to$0.49).}
\label{fig:supp_trails}
\end{figure*}
\subsection{Quality of the Label-Free Confusions}
\label{sec:supp_em}

\paragraph{When no observation names the answer.} On 17.0\%, 11.1\% and 7.5\% of the TCGA, GTEx and PANDA slides, no view votes for the true answer, so the bounded evidence (a) has nothing to support it. The confusion matrix reads these votes as typical misreadings of the true tissue (\cref{fig:confusion}): on these slides, \method is correct on 38.9\%, 44.7\% and 32.0\%, against 23.0\%, 17.3\% and 14.0\% with (a) alone.

\paragraph{Agreement with the true confusions.} Estimated without labels, $B$ tracks the true confusions of the votes on held-out slides: their off-diagonal entries correlate with $r=0.78$ on TCGA and $0.75$ on GTEx (\cref{tab:supp_bq}).

\input{sec/X_tab_bq}

\subsection{Repeated Views}
\label{sec:supp_copies}
Repeated views of similar tissue should not change the answer. We copy one observation 16 times; the copies add no information, so ideally no answer changes. \method changes far fewer answers than naive Bayes and LLM aggregation (\cref{tab:copies}).

\begin{table}[H]
\centering
\footnotesize
\setlength{\tabcolsep}{4pt}
\tabno{20}\caption{\textbf{Repeated views.} Share of answers that change when one observation is copied 16 times (slide classification, \%); $\downarrow$: lower is better. Mean of three seeds.}
\label{tab:copies}
\vspace{-0.6em}
\begin{tabular}{@{}lccc@{}}
\toprule
Rule & TCGA\,$\downarrow$ & GTEx\,$\downarrow$ & PANDA\,$\downarrow$ \\
\midrule
LLM aggregation & 15.5 & 8.6 & 27.0 \\
Naive Bayes & 59.4 & 30.7 & 48.6 \\
\rowcolor{cbblue!7}\method & \textbf{8.9} & \textbf{5.8} & \textbf{6.1} \\
\bottomrule
\end{tabular}
\vspace{-0.8em}
\end{table}

\subsection{Calibration}
\label{sec:supp_calib}
On the same evidence, naive Bayes is nearly certain but correct far less often, while \method is well calibrated on every task (\cref{tab:calib}).

\begin{table}[H]
\centering
\footnotesize
\setlength{\tabcolsep}{3.5pt}
\tabno{21}\caption{\textbf{Calibration} of the final belief on the same evidence; $\downarrow$: lower is better. ECE: ten bins of the confidence of the answer; Brier: squared error of the full belief. Mean of three seeds. EVQA: ExpertVQA; SB: SlideBench.}
\label{tab:calib}
\begin{tabular}{@{}llccccc@{}}
\toprule
& & TCGA & GTEx & PANDA & EVQA & SB \\
\midrule
\multirow{2}{*}{ECE$\downarrow$} & Naive Bayes & 0.527 & 0.331 & 0.671 & 0.545 & 0.436 \\
 & \cellcolor{cbblue!7}\method & \cellcolor{cbblue!7}\textbf{0.103} & \cellcolor{cbblue!7}\textbf{0.066} & \cellcolor{cbblue!7}\textbf{0.127} & \cellcolor{cbblue!7}\textbf{0.116} & \cellcolor{cbblue!7}\textbf{0.088} \\
\midrule
\multirow{2}{*}{Brier$\downarrow$} & Naive Bayes & 1.09 & 0.67 & 1.40 & 1.09 & 0.87 \\
 & \cellcolor{cbblue!7}\method & \cellcolor{cbblue!7}\textbf{0.44} & \cellcolor{cbblue!7}\textbf{0.28} & \cellcolor{cbblue!7}\textbf{0.85} & \cellcolor{cbblue!7}\textbf{0.66} & \cellcolor{cbblue!7}\textbf{0.55} \\
\bottomrule
\end{tabular}
\end{table}
\section{Integrating Pathology Foundation Models}
\label{sec:supp_titan}
\label{sec:titan}

\subsection{A Stronger Encoder}
\label{sec:supp_stronger}
\Cref{tab:supp_models} gives \cref{tab:stronger} with standard deviations.

\subsection{Other Observers}
\label{sec:supp_swap}
With every Observer, \method stays far more accurate than both agents (\cref{tab:supp_swap}).
\begin{table}[t]
\centering
\footnotesize
\setlength{\tabcolsep}{2.6pt}
\tabno{22}\caption{\textbf{Other Observers} on slide classification (BACC, \%): the Observer of all three agents is replaced, without training. Mean of three seeds.}
\label{tab:supp_swap}
\begin{tabular}{@{}lcccc>{\columncolor{cbblue!7}}c>{\columncolor{cbblue!7}}c@{}}
\toprule
 & \multicolumn{2}{c}{PathAgent} & \multicolumn{2}{c}{PathNavigate} & \multicolumn{2}{c}{\method} \\
\cmidrule(lr){2-3}\cmidrule(lr){4-5}\cmidrule(l){6-7}
Observer & TCGA & GTEx & TCGA & GTEx & TCGA & GTEx \\
\midrule
Patho-R1-7B (default) & 14.5 & 45.0 & 6.2 & 33.3 & \textbf{75.3} & \textbf{83.4} \\
Quilt-LLaVA~\cite{quiltllava} & 6.8 & 33.4 & 4.4 & 11.9 & \textbf{71.5} & \textbf{84.0} \\
PathGen-LLaVA~\cite{pathgen} & 13.0 & 53.3 & 6.8 & 41.5 & \textbf{72.1} & \textbf{84.9} \\
\bottomrule
\end{tabular}
\end{table}
\begin{table}[t]
\centering
\footnotesize
\setlength{\tabcolsep}{3pt}
\tabno{23}\caption{\textbf{A stronger encoder:} \cref{tab:stronger} with mean $\pm$ std of three seeds (\%). EVQA: ExpertVQA; SB: SlideBench.}
\label{tab:supp_models}
\resizebox{\linewidth}{!}{%
\begin{tabular}{@{}lccccc@{}}
\toprule
Method & TCGA & GTEx & PANDA & EVQA & SB \\
\midrule
PathAgent & \pms{15.7}{2.5} & \pms{43.4}{5.5} & \pms{22.1}{1.5} & \pms{41.9}{2.1} & \pms{54.5}{1.5} \\
PathNavigate & \pms{6.3}{1.5} & \pms{30.8}{3.0} & \pms{18.8}{0.9} & \pms{47.1}{0.5} & \pms{58.0}{0.6} \\
\rowcolor{cbblue!7}\method & \textbf{\pms{78.7}{1.0}} & \textbf{\pms{86.4}{0.3}} & \textbf{\pms{25.7}{0.6}} & \textbf{\pms{48.2}{5.0}} & \textbf{\pms{58.9}{2.0}} \\
\bottomrule
\end{tabular}}
\end{table}

\section{Qualitative Evaluation}
\label{sec:supp_trails}
On two more slides, \method weighs each view by its reliability and answers correctly (\cref{fig:supp_trails}).

%% file: sec/X_tab_stats.tex
\begin{table}[t]
\centering
\footnotesize
\setlength{\tabcolsep}{2.5pt}
\tabno{7}\caption{\textbf{Paired comparisons} on the same slides. $\Delta$: gain of \method in BACC (classification) or reduction in AURC (slide VQA, last two rows; lower is better, \cref{tab:trust}), computed before rounding, so it can differ by 0.1 from the difference of rounded entries. CI: patient-level bootstrap 95\% interval (2,000 resamples). $p$: largest exact McNemar $p$ of three seeds (classification only).}
\label{tab:supp_stats}
\resizebox{\linewidth}{!}{%
\begin{tabular}{@{}lccc@{}}
\toprule
Comparison & $\Delta$ & 95\% CI & McNemar $p$ \\
\midrule
TCGA: \method $-$ LLM aggregation & +50.5 & [46.7, 54.4] & 1.3e-15 \\
TCGA: \method $-$ naive Bayes & +39.3 & [35.2, 43.7] & 1.1e-10 \\
GTEx: \method $-$ LLM aggregation & +28.3 & [24.1, 32.7] & 2.2e-10 \\
GTEx: \method $-$ naive Bayes & +17.5 & [13.6, 21.9] & 2.2e-07 \\
\midrule
TCGA: \method $-$ PathAgent & +60.8 & [56.7, 65.0] & 5.7e-23 \\
TCGA: \method $-$ PathNavigate & +69.1 & [64.9, 73.2] & 8.6e-35 \\
GTEx: \method $-$ PathAgent & +38.5 & [33.2, 44.1] & 1.5e-12 \\
GTEx: \method $-$ PathNavigate & +50.2 & [45.6, 54.5] & 4.3e-21 \\
\midrule
ExpertVQA: AURC, naive Bayes $-$ \method & +0.087 & [0.020, 0.154] & -- \\
SlideBench: AURC, naive Bayes $-$ \method & +0.100 & [0.041, 0.158] & -- \\
\bottomrule
\end{tabular}}
\end{table}

%% file: sec/X_tab_same.tex
\begin{table}[t]
\centering
\footnotesize
\setlength{\tabcolsep}{2.2pt}
\tabno{9}\caption{\textbf{Same-evidence aggregation on TCGA,} extending \cref{tab:memory}. Mean of three seeds. Conf.: mean confidence on correct (\ding{51}) and wrong (\ding{55}) answers.}
\label{tab:supp_same_tcga}
\vspace{-0.6em}
\begin{tabular}{@{}lccccccc@{}}
\toprule
\multirow{2}{*}{Aggregation} & \multirow{2}{*}{Acc\,$\uparrow$} & \multirow{2}{*}{BACC\,$\uparrow$} & \multirow{2}{*}{ECE\,$\downarrow$} & \multirow{2}{*}{Brier\,$\downarrow$} & \multirow{2}{*}{NLL\,$\downarrow$} & \multicolumn{2}{c}{Conf.} \\
\cmidrule(l){7-8}
 & & & & & & \ding{51} & \ding{55}\,$\downarrow$ \\
\midrule
LLM aggregation & 33.9 & 24.8 & -- & -- & -- & -- & -- \\
Raw product & 30.9 & 20.9 & 0.687 & 1.38 & 9.30 & 99.6 & 99.6 \\
Naive Bayes & 41.6 & 36.0 & 0.527 & 1.09 & 7.09 & 98.0 & 91.7 \\
\rowcolor{cbblue!7}\method & \textbf{69.8} & \textbf{75.3} & \textbf{0.103} & \textbf{0.44} & \textbf{1.16} & 68.5 & \textbf{41.9} \\
\bottomrule
\end{tabular}
\end{table}

\begin{table}[t]
\centering
\footnotesize
\setlength{\tabcolsep}{2.2pt}
\tabno{10}\caption{\textbf{Same-evidence aggregation on GTEx,} extending \cref{tab:memory}. Mean of three seeds. Conf.: mean confidence on correct (\ding{51}) and wrong (\ding{55}) answers.}
\label{tab:supp_same_gtex}
\vspace{-0.6em}
\begin{tabular}{@{}lccccccc@{}}
\toprule
\multirow{2}{*}{Aggregation} & \multirow{2}{*}{Acc\,$\uparrow$} & \multirow{2}{*}{BACC\,$\uparrow$} & \multirow{2}{*}{ECE\,$\downarrow$} & \multirow{2}{*}{Brier\,$\downarrow$} & \multirow{2}{*}{NLL\,$\downarrow$} & \multicolumn{2}{c}{Conf.} \\
\cmidrule(l){7-8}
 & & & & & & \ding{51} & \ding{55}\,$\downarrow$ \\
\midrule
LLM aggregation & 56.3 & 55.1 & -- & -- & -- & -- & -- \\
Raw product & 56.0 & 54.6 & 0.429 & 0.86 & 5.82 & 99.3 & 97.7 \\
Naive Bayes & 65.8 & 65.9 & 0.331 & 0.67 & 4.23 & 99.4 & 97.5 \\
\rowcolor{cbblue!7}\method & \textbf{81.2} & \textbf{83.4} & \textbf{0.066} & \textbf{0.28} & \textbf{0.85} & 90.3 & \textbf{60.6} \\
\bottomrule
\end{tabular}
\end{table}

%% file: sec/X_tab_br.tex
\begin{table}[H]
\centering
\footnotesize
\setlength{\tabcolsep}{4pt}
\tabno{18}\caption{\textbf{Label-free statistics from other slides} (BACC, \%). The answer biases and $B$ are estimated from the other half of the test slides (default) or only from TCGA slides of the slide VQA tasks, which do not overlap with the classification set. GTEx has no such slides. Mean of three seeds.}
\label{tab:supp_baserate}
\begin{tabular}{@{}lcc@{}}
\toprule
Estimated from & TCGA & GTEx \\
\midrule
\rowcolor{cbblue!7}Other half of the test slides (\method) & \textbf{75.3} & \textbf{83.4} \\
253 slides of the slide VQA tasks & 74.2 & -- \\
\bottomrule
\end{tabular}
\end{table}

%% file: sec/X_tab_bq.tex
\begin{table}[H]
\centering
\footnotesize
\setlength{\tabcolsep}{2.5pt}
\tabno{19}\caption{\textbf{Label-free against true confusions.} $B$, estimated without labels on one half of the slides, against the true confusions of the votes on the other half (labels used for this check only). $r$: Pearson correlation of off-diagonal entries; cos: cosine and KL: KL divergence of the rows.}
\label{tab:supp_bq}
\begin{tabular}{@{}lccc@{}}
\toprule
Task & $r\uparrow$ & cos$\uparrow$ & KL$\downarrow$ \\
\midrule
TCGA & 0.78 & 0.83 & 0.27 \\
GTEx & 0.75 & 0.91 & 0.18 \\
\bottomrule
\end{tabular}
\end{table}